\documentclass[12pt]{article}
\usepackage[margin=1in]{geometry}

\usepackage[hyphens]{xurl}
\usepackage[
    backend=biber,
    style=nature,
    sorting=none,
    natbib=true,
    url=true, 
    doi=true,
    eprint=false,
    maxbibnames=99,
    date=year,
    datamodel=protocollessurl,
]{biblatex}
\AtEveryBibitem{\clearlist{language}}
\AtEveryBibitem{\clearfield{issn}}
\AtEveryCitekey{\clearfield{issn}}

\DeclareSourcemap{
  \maps{
    \map{
      \pertype{inproceedings}
      \step[fieldset=editor, null]
      \step[fieldset=publisher, null]
    }
  }
}

\DeclareFieldFormat{volume}{#1} %
\DeclareSourcemap{
  \maps[datatype=bibtex]{
    \map{
      \step[fieldsource=url, final=true]
      \step[fieldset=protocollessurl, origfieldval, final=true]
      \step[fieldsource=protocollessurl, match=\regexp{\A(ht|f)tp(s)?:\/\/}, replace={}]
    }
  }
}

\DeclareFieldFormat{url}{%
  \mkbibacro{URL}\addcolon\space
  \href{#1}{\nolinkurl{\thefield{protocollessurl}}}}

\usepackage{times}

\usepackage{graphicx} %
\usepackage{booktabs}       %
\usepackage{tabularx}
\usepackage[hidelinks,breaklinks]{hyperref}       %
\usepackage{bm} %
\usepackage{mathtools} %
\usepackage{sidecap} %
\usepackage{adjustbox}
\usepackage{amsfonts} %
\usepackage[group-separator={,}]{siunitx} %
\usepackage{nicefrac}       %
\usepackage{placeins} %
\usepackage{xcolor}
\usepackage[font={small}]{caption}
\usepackage[font={sf,normalsize}]{subcaption} %
\usepackage{float} %
\usepackage{tabularx}
\usepackage{sidecap}

\newenvironment{sciabstract}{%
\begin{quote} \bf}
{\end{quote}}

\title{\vspace{-2cm}Testing the limits of natural language models for predicting human language judgments}

\author
{Tal Golan$^{1,2\ast\dagger}$, Matthew Siegelman$^{3\ast}$,\\Nikolaus Kriegeskorte$^{1,3,4,5}$, Christopher Baldassano$^{3}$\\
\\
\normalsize{$^{1}$Zuckerman Mind Brain Behavior Institute, Columbia University,}
\normalsize{New York, NY, USA}\\
\normalsize{$^{2}$Department of Cognitive and Brain Sciences, Ben-Gurion University of the Negev,}
\normalsize{Be'er-Sheva, Israel}\\
\normalsize{$^{3}$Department of Psychology, Columbia University,}
\normalsize{New York, NY, USA}\\
\normalsize{$^{4}$Department of Neuroscience, Columbia University,}
\normalsize{New York, NY, USA}\\
\normalsize{$^{5}$Department of Electrical Engineering, Columbia University,}
\normalsize{New York, NY, USA}\\
\\
\normalsize{$^\ast$The first two authors contributed equally to this work.}\\
\normalsize{$^\dagger$To whom correspondence should be addressed; E-mail: golan.neuro@bgu.ac.il}
}

\date{}

\DeclareMathOperator*{\argmin}{argmin} %
\DeclareMathOperator*{\sign}{sign}

\newcommand{\beginextdata}{%
        \setcounter{table}{0}
        \setcounter{figure}{0}
     }

\newcommand{\beginsupplement}{%
        \setcounter{table}{0}
        \renewcommand{\thetable}{S\arabic{table}}%
        \setcounter{figure}{0}
        \renewcommand{\thefigure}{S\arabic{figure}}%
     }

\providecommand{\keywords}[1]
{
  \small	
  \textbf{\textit{Keywords---}} #1
}

\begin{document} 

\sisetup{detect-weight=true, detect-family=true} %

\maketitle

\vspace{-18pt}
\begin{sciabstract} %
Neural network language models appear to be increasingly aligned with how humans process and generate language, but identifying their weaknesses through adversarial examples is challenging due to the discrete nature of language and the complexity of human language perception. We bypass these limitations by turning the models against each other. We generate \emph{controversial sentence pairs} for which two language models disagree about which sentence is more likely to occur. Considering nine language models (including n-gram, recurrent neural networks, and transformers), we created hundreds of controversial sentence pairs through synthetic optimization or by selecting sentences from a corpus. Controversial sentence pairs proved highly effective at revealing model failures and identifying models that aligned most closely with human judgments of which sentence is more likely. The most human-consistent model tested was GPT-2, although experiments also revealed significant shortcomings of its alignment with human perception.
\end{sciabstract}

\maketitle

\section{Introduction}\label{sec1}

Neural network language models are not only key tools in natural language processing (NLP) but are also drawing an increasing scientific interest as potential models of human language-processing. Ranging from recurrent neural networks \cite{rumelhart1986learning, hochreiter1997long} to transformers \cite{devlin2019bert, liu2019roberta, Conneau2019Cross, clark2020electra, radford2019language}, each of these language models (explicitly or implicitly) defines a probability distribution over strings of words, predicting which sequences are likely to occur in natural language. There is substantial evidence from measures such as reading times \cite{goodkind2018predictive}, functional MRI \cite{shain2020fmri}, scalp EEG \cite{broderick2018electrophysiological}, and intracranial ECoG \cite{goldstein_shared_2022} that humans are sensitive to the relative probabilities of words and sentences as captured by language models, even among sentences that are grammatically correct and semantically meaningful. Furthermore, model-derived sentence probabilities can also predict human graded acceptability judgments \cite{lau_2017_grammaticality, lau_2020_how}. These successes, however, have not yet addressed two central questions of interest: (1) Which of the models is best-aligned with human language processing? (2) How close is the best-aligned model to the goal of fully capturing human judgments?

A predominant approach for evaluating language models is to use a set of standardized benchmarks such as those in the General Language Understanding Evaluation (GLUE) \cite{wang2019glue}, or its successor, SuperGLUE \cite{wang2019superglue}. Though instrumental in evaluating the utility of language models for downstream NLP tasks, these benchmarks prove insufficient for comparing such models as candidate explanations of human language-processing. Many components of these benchmarks do not aim to measure human alignment but rather the usefulness of the models' language representation when tuned to a specific downstream task. Some benchmarks challenge language models more directly by comparing the probabilities they assign to grammatical and ungrammatical sentences (e.g., BLiMP \cite{Warstadt2020BLiMP}). However, since such benchmarks are driven by theoretical linguistic considerations, they might fail to detect novel, unexpected ways in which language models may diverge from human language understanding. Lastly, an additional practical concern is that the rapid pace of NLP research has led to quick saturation of these kinds of static benchmarks, making it difficult to distinguish between models \cite{kiela2021dynabench}.

One proposed solution to these issues is the use of dynamic human-in-the-loop benchmarks in which people actively stress-test models with an evolving set of tests. However, this approach faces the major obstacle that ``finding interesting examples is rapidly becoming a less trivial task'' \cite{kiela2021dynabench}. We propose to complement human-curated benchmarks with model-driven evaluation. Guided by model predictions rather than experimenter intuitions, we would like to identify particularly informative test sentences, where different models make divergent predictions. This approach of running experiments mathematically optimized to ``put in jeopardy'' particular models belongs to a long-standing scientific philosophy of design optimization \cite{box_hill}. We can find these critical sentences in large corpora of natural language or synthesize novel test sentences that reveal how different models generalize beyond their training distributions.

We propose here a systematic, model-driven approach for comparing language models in terms of their consistency with human judgments. We generate \emph{controversial sentence pairs}: pairs of sentences designed such that two language models strongly disagree about which sentence is more likely to occur. In each of these sentence pairs, one model assigns a higher probability to the first sentence than the second sentence, while the other model prefers the second sentence to the first. We then collect human judgments of which sentence in each pair is more probable to settle this dispute between the two models.

This approach builds on previous work on controversial images for models of visual classification \cite{golan_controversial_2020}. That work relied on absolute judgments of a single stimulus, which are appropriate for classification responses. However, asking the participants to rate each sentence's probability on an absolute scale is complicated by between-trial context effects common in magnitude estimation tasks \cite{cross_sequential_1973,foley_pervasiveness_1990,petzschner_bayesian_2015}, which have been shown to impact judgments like acceptability \cite{Greenbaum1977-sp}. A binary forced-choice behavioral task presenting the participants with a choice between two sentences in each trial, the approach we used here, minimizes the role of between-trial context effects by setting an explicit local context within each trial. Such an approach has been previously used for measuring sentence acceptability \cite{schütze_sprouse_2014} and provides substantially more statistical power compared to designs in which acceptability ratings are provided for single sentences \cite{Sprouse2017-mu}.

Our experiments demonstrate that (1) it is possible to procedurally generate controversial sentence pairs for all common classes of language models, either by selecting pairs of sentences from a corpus or by iteratively modifying natural sentences to yield controversial predictions; (2) the resulting controversial sentence pairs enable efficient model comparison between models that otherwise are seemingly equivalent in their human consistency; and (3) all current NLP model classes incorrectly assign high probability to some non-natural sentences (one can modify a natural sentence such that its model probability does not decrease but human observers reject the sentence as unnatural). This framework for model comparison and model testing can give us new insight into the classes of models that best align with human language perception and suggest directions for future model development.

\section{Results}\label{sec2}

We acquired judgments from 100 native English speakers tested online. In each experimental trial, the participants were asked to judge which of two sentences they would be ``more likely to encounter in the world, as either speech or written text'', and provided a rating of their confidence in their answer on a 3-point scale (see Extended Data Fig.~\ref{fig:extfig_1_example_trial} for a trial example). The experiment was designed to compare nine different language models (Supplementary Section \ref{sup_methods_language_models}): probability models based on corpus frequencies of 2-word and 3-word sequences (2-grams and 3-grams) and a range of neural network models comprising a recurrent neural network (RNN), a long short-term memory network (LSTM), and five transformer models (BERT, RoBERTa, XLM, ELECTRA, and GPT-2).

\subsection{Efficient model comparison using natural controversial pairs}

\begin{figure}[htbp]%
	\centering
	\includegraphics{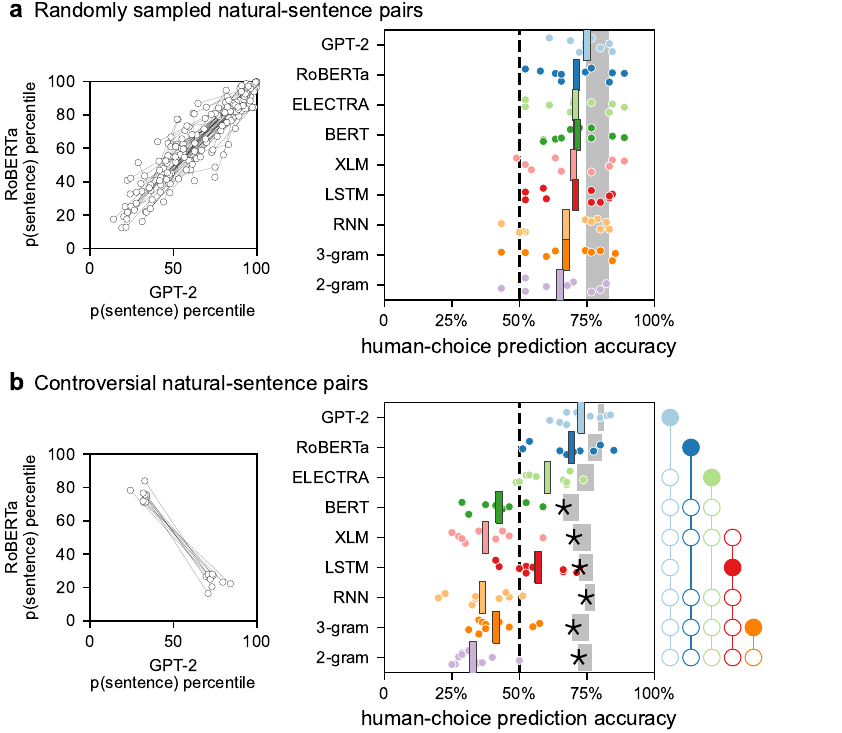}
	\caption{\textbf{Model comparison using natural sentences.} \textbf{(a)} (Left) Percentile-transformed sentence probabilities for GPT-2 and RoBERTa (defined relative to all sentences used in the experiment) for randomly-sampled pairs of natural sentences. Each pair of connected dots depicts one sentence pair. The two models are highly congruent in their rankings of sentences within a pair (lines have upward slope). (Right) Accuracy of model predictions of human choices, measured as the proportion of trials in which the same sentence was preferred by both the model and the human participant. Each dot depicts the prediction accuracy of one candidate model averaged across a group of 10 participants presented with a unique set of trials. The colored bars depict grand-averages across all 100 participants. The gray bar is the noise ceiling whose left and right edges are lower and upper bounds on the grand-average performance an ideal model would achieve (based on the consistency across human subjects). There were no significant differences in model performance on the randomly sampled natural sentences. \textbf{(b)} (Left) Controversial natural-sentence pairs were selected such that the models' sentence probability ranks were incongruent (lines have downward slope). (Right) Controversial sentence pairs enable efficient model comparison, revealing that BERT, XLM, LSTM, RNN and the n-gram models perform significantly below the noise ceiling (asterisks indicate significance---two-sided Wilcoxon signed-rank test, controlling the false discovery rate for nine comparisons at \textit{q}~$<$~.05). On the right of the plot, each closed circle indicates a model significantly dominating alternative models indicated by open circles (two-sided Wilcoxon signed-rank test, controlling the false discovery rate for all 36 model pairs at \textit{q}~$<$~.05). GPT-2 outperforms all models except RoBERTA at predicting human judgments.}
    \label{fig:fig1_natural_binarized_accuracy}
\end{figure}

As a baseline, we randomly sampled and paired 8-word sentences from a corpus of Reddit comments. However, as shown in Fig.~\ref{fig:fig1_natural_binarized_accuracy}a, these sentences fail to uncover meaningful differences between the models. For each sentence pair, all models tend to prefer the same sentence (Extended Data Fig.~\ref{fig:extfig_2_randomly_sampled_sents_model_agreement}), and therefore perform similarly in predicting human preference ratings (see Supplementary Section \ref{sup_results_randomly_sampled}).

Instead, we can use an optimization procedure (Supplementary Section \ref{sup_methods_selection}) to search for \textit{controversial} sentence pairs, in which one language model assigns a high probability (above the median probability for natural sentences) only to sentence 1 and a second language model assigns a high probability only to sentence 2; see examples in Table~\ref{tab:tab1_natural_controversial_sentence_pairs}. Measuring each model's accuracy in predicting human choices for sentence pairs in which it was one of the two targeted models indicated many significant differences in terms of model-human alignment (Fig.~\ref{fig:fig1_natural_binarized_accuracy}b), with GPT-2 and RoBERTa showing the best human consistency and 2-gram the worst. We can also compare each model pair separately (using only the stimuli targeting that model pair), yielding a similar pattern of pairwise dominance (Extended Data Fig.~\ref{fig:extfig_3_pairwise_model}a). All models except GPT-2, RoBERTa, and ELECTRA performed significantly below our lower bound on the noise ceiling (the accuracy obtained by predicting each participant's responses from the other participants' responses), indicating a misalignment between these models' predictions and human judgments which was only revealed when using controversial sentence pairs.

\begin{table}[t]
\centering
\scriptsize
 
\begin{adjustbox}{center}
\begin{tabularx}{18cm}{lllc}
        \toprule
                                                               sentence &                               log probability (model 1) &                               log probability (model 2) &   \# human choices \\
        \midrule
                      $n_1$: Rust is generally caused by salt and sand. &             $\log p(n_1 | \textrm{GPT-2})=$\num{-50.72} &  $\log p(n_1 | \textrm{ELECTRA})=$\textbf{\num{-38.54}} &  \textbf{\num{10}} \\
                        $n_2$: Where is Vernon Roche when you need him. &    $\log p(n_2 | \textrm{GPT-2})=$\textbf{\num{-32.26}} &           $\log p(n_2 | \textrm{ELECTRA})=$\num{-58.26} &            \num{0} \\\midrule
         $n_1$: Excellent draw and an overall great smoking experience. &           $\log p(n_1 | \textrm{RoBERTa})=$\num{-67.78} &    $\log p(n_1 | \textrm{GPT-2})=$\textbf{\num{-36.76}} &  \textbf{\num{10}} \\
                       $n_2$: I should be higher and tied to inflation. &  $\log p(n_2 | \textrm{RoBERTa})=$\textbf{\num{-54.61}} &             $\log p(n_2 | \textrm{GPT-2})=$\num{-50.31} &            \num{0} \\\midrule
                             $n_1$: You may try and ask on their forum. &           $\log p(n_1 | \textrm{ELECTRA})=$\num{-51.44} &     $\log p(n_1 | \textrm{LSTM})=$\textbf{\num{-44.24}} &  \textbf{\num{10}} \\
                    $n_2$: I love how they look like octopus tentacles. &  $\log p(n_2 | \textrm{ELECTRA})=$\textbf{\num{-35.51}} &              $\log p(n_2 | \textrm{LSTM})=$\num{-66.66} &            \num{0} \\\midrule
            $n_1$: Grow up and quit whining about minor inconveniences. &              $\log p(n_1 | \textrm{BERT})=$\num{-82.74} &    $\log p(n_1 | \textrm{GPT-2})=$\textbf{\num{-35.66}} &  \textbf{\num{10}} \\
              $n_2$: The extra a is the correct Sanskrit pronunciation. &     $\log p(n_2 | \textrm{BERT})=$\textbf{\num{-51.06}} &             $\log p(n_2 | \textrm{GPT-2})=$\num{-51.10} &            \num{0} \\\midrule
                     $n_1$: I like my password manager for this reason. &               $\log p(n_1 | \textrm{XLM})=$\num{-68.93} &  $\log p(n_1 | \textrm{RoBERTa})=$\textbf{\num{-49.61}} &  \textbf{\num{10}} \\
                             $n_2$: Kind of like clan of the cave bear. &      $\log p(n_2 | \textrm{XLM})=$\textbf{\num{-44.24}} &           $\log p(n_2 | \textrm{RoBERTa})=$\num{-67.00} &            \num{0} \\\midrule
                  $n_1$: We have raised a Generation of Computer geeks. &              $\log p(n_1 | \textrm{LSTM})=$\num{-66.41} &  $\log p(n_1 | \textrm{ELECTRA})=$\textbf{\num{-36.57}} &  \textbf{\num{10}} \\
                         $n_2$: I mean when the refs are being sketchy. &     $\log p(n_2 | \textrm{LSTM})=$\textbf{\num{-42.04}} &           $\log p(n_2 | \textrm{ELECTRA})=$\num{-52.28} &            \num{0} \\\midrule
               $n_1$: This is getting ridiculous and ruining the hobby. &              $\log p(n_1 | \textrm{RNN})=$\num{-100.65} &     $\log p(n_1 | \textrm{LSTM})=$\textbf{\num{-43.50}} &  \textbf{\num{10}} \\
                     $n_2$: I think the boys and invincible are better. &      $\log p(n_2 | \textrm{RNN})=$\textbf{\num{-45.16}} &              $\log p(n_2 | \textrm{LSTM})=$\num{-59.00} &            \num{0} \\\midrule
                 $n_1$: Then attach them with the supplied wood screws. &           $\log p(n_1 | \textrm{3-gram})=$\num{-119.09} &    $\log p(n_1 | \textrm{GPT-2})=$\textbf{\num{-34.84}} &  \textbf{\num{10}} \\
                           $n_2$: Sounds like you were used both a dog. &   $\log p(n_2 | \textrm{3-gram})=$\textbf{\num{-92.07}} &             $\log p(n_2 | \textrm{GPT-2})=$\num{-52.84} &            \num{0} \\\midrule
                   $n_1$: Cream cheese with ham and onions on crackers. &           $\log p(n_1 | \textrm{2-gram})=$\num{-131.99} &  $\log p(n_1 | \textrm{RoBERTa})=$\textbf{\num{-54.62}} &  \textbf{\num{10}} \\
                   $n_2$: I may have to parallel process that drinking. &  $\log p(n_2 | \textrm{2-gram})=$\textbf{\num{-109.46}} &           $\log p(n_2 | \textrm{RoBERTa})=$\num{-70.69} &            \num{0} \\
        \bottomrule
        \end{tabularx}
        \end{adjustbox}
        \scriptsize
    \caption{\textbf{Examples of controversial natural-sentence pairs that maximally contributed to each model's prediction error.} For each model (double row, ``model 1''), the table shows results for two sentences on which the model failed severely. In each case, the failing model 1 prefers sentence $n_2$ (higher log probability bolded), while the model it was pitted against (``model 2'') and all 10 human subjects presented with that sentence pair prefer sentence $n_1$. (When more than one sentence pair induced an equal maximal error in a model, the example included in the table was chosen at random.)\label{tab:tab1_natural_controversial_sentence_pairs}}
\end{table}

\FloatBarrier

\subsection{Greater model disentanglement with synthetic sentence pairs}

\begin{figure}[ht]
	\centering
	\includegraphics{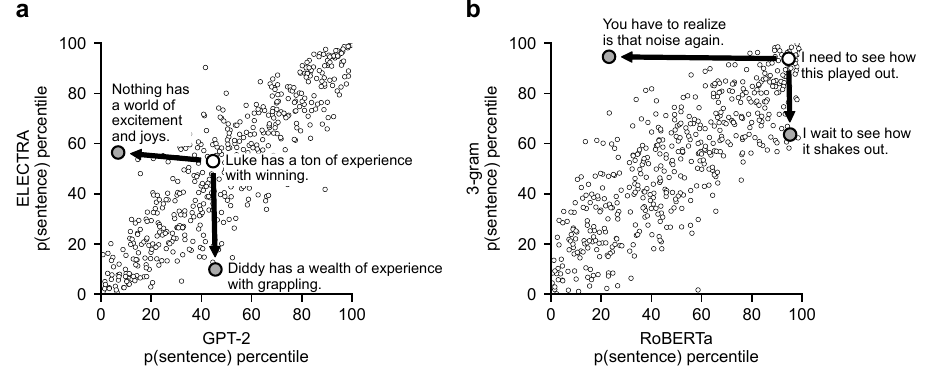}
	\caption{\textbf{Synthesizing controversial sentence pairs.} The small open dots denote 500 randomly sampled natural sentences. The big open dot denotes the natural sentence used for initializing the controversial sentence optimization, and the closed dots are the resulting synthetic sentences. \textbf{(a)} In this example, we start with the randomly sampled natural sentence ``Luke has a ton of experience with winning''. If we adjust this sentence to minimize its probability according to GPT-2 (while keeping the sentence at least as likely as the natural sentence according to ELECTRA), we obtain the synthetic sentence ``Nothing has a world of excitement and joys''. By repeating this procedure while switching the roles of the models, we generate the synthetic sentence ``Diddy has a wealth of experience with grappling'', which decreases ELECTRA's probability while slightly increasing GPT-2's. \textbf{(b)} In this example, we start with the randomly sampled natural sentence ``I need to see how this played out''. If we adjust this sentence to minimize its probability according to RoBERTa (while keeping the sentence at least as likely as the natural sentence according to 3-gram), we obtain the synthetic sentence ``You have to realize is that noise again''. If we instead decrease only 3-gram's probability, we generate the synthetic sentence ``I wait to see how it shakes out''.
	\label{fig:fig2_sentence_optimization_illustration}}
\end{figure}

Selecting controversial natural-sentence pairs may provide greater power than randomly sampling natural-sentence pairs, but this search procedure considers a very limited part of the space of possible sentence pairs. Instead, we can iteratively replace words in a natural sentence to drive different models to make opposing predictions, forming \textit{synthetic} controversial sentences that may lay outside any natural language corpora, as illustrated in Fig.~\ref{fig:fig2_sentence_optimization_illustration} (see Methods, ``\nameref{ssec:generating_synthetic}'' for full details). Examples of controversial synthetic-sentence pairs that maximally contributed to the models' prediction error appear in Table~\ref{tab:tab2_synthetic_controversial_sentence_pairs}.

We evaluated how well each model predicted the human sentence choices in all of the controversial synthetic-sentence pairs in which the model was one of the two models targeted (Fig.~\ref{fig:fig3_synthetic_binarized}a). This evaluation of model-human alignment resulted in an even greater separation between the models' prediction accuracies than was obtained when using controversial natural-sentence pairs, pushing the weaker models (RNN, 3-gram, and 2-gram) far below the 50\% chance accuracy level. GPT-2, RoBERTa, and ELECTRA were found to be significantly more accurate than the alternative models (BERT, XLM, LSTM, RNN, 3-gram, and 2-gram) in predicting the human responses to these trials (with similar results when comparing model pair separately, see Extended Data Fig.~\ref{fig:extfig_3_pairwise_model}b). All of the models except for GPT-2 were found to be significantly below the lower bound on the noise ceiling, demonstrating misalignment with human judgments.

\subsection{Pairs of natural and synthetic sentences uncover blindspots}

\begin{figure}[t]
	\centering
	\includegraphics{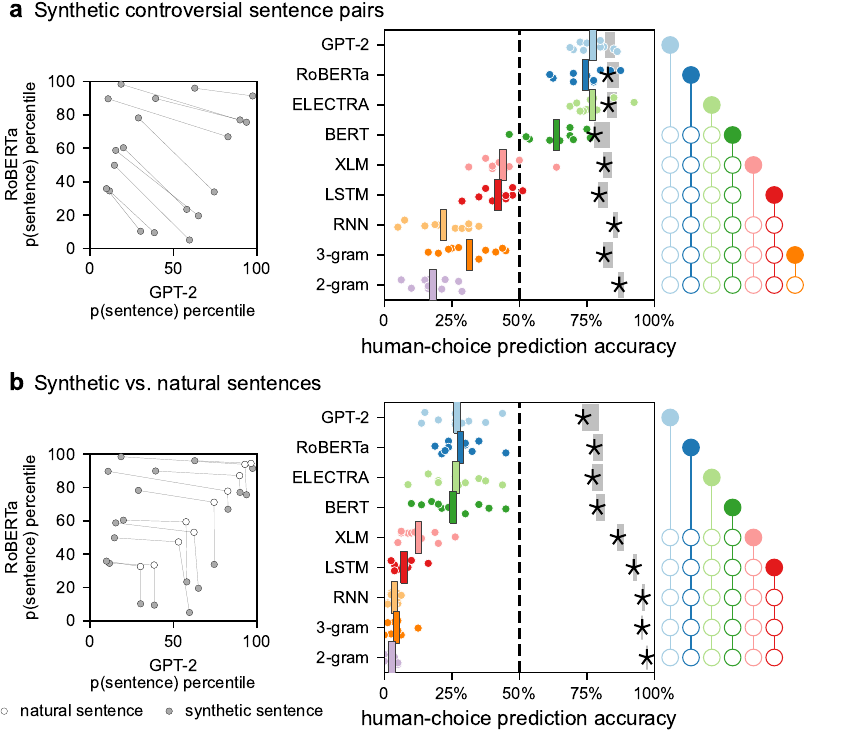}
	\caption{\textbf{Model comparison using synthetic sentences.} \textbf{(a)} (Left) Percentile-transformed sentence probabilities for GPT-2 and RoBERTa for controversial synthetic-sentence pairs. Each pair of connected dots depict one sentence pair. (Right) Model prediction accuracy, measured as the proportion of trials in which the same sentence was preferred by both the model and the human participant. GPT-2, RoBERTa and ELECTRA significantly outperformed the other models (two-sided Wilcoxon signed-rank test, controlling the false discovery rate for all 36 model comparisons at \textit{q}~$<$~.05). All of the models except for GPT-2 were found to perform below the noise ceiling (gray) of predicting each participant's choices from the majority votes of the other participants (asterisks indicate significance---two-sided Wilcoxon signed-rank test, controlling the false discovery rate for nine comparisons at \textit{q}~$<$~.05). \textbf{(b)} (Left) Each connected triplet of dots depicts a natural sentence and its derived synthetic sentences, optimized to decrease the probability only under GPT-2 (left dots in a triplet) or only under RoBERTa (bottom dots in a triplet). (Right) Each model was evaluated across all of the synthetic-natural sentence pairs for which it was targeted to keep the synthetic sentence at least as probable as the natural sentence (see Extended Data Fig.~\ref{fig:extfig_6_synthetic_binarized_alternative_trial_selection} for the complementary data binning). This evaluation yielded a below-chance prediction accuracy for all of the models, which was also significantly below the lower bound on the noise ceiling. This indicates that, although the models assessed that these synthetic sentences were at least as probable as the original natural sentence, humans disagreed and showed a systematic preference for the natural sentence. See Fig.~\ref{fig:fig1_natural_binarized_accuracy}'s caption for details on the visualization conventions used in this figure.
    \label{fig:fig3_synthetic_binarized}}
\end{figure}

Last, we considered trials in which the participants were asked to choose between a natural sentence and one of the synthetic sentences which was generated from that natural sentence. If the language model is fully aligned with human judgments, we would expect humans to agree with the model, and select the synthetic sentence at least as much as the natural sentence. In reality, human participants showed a systematic preference for the natural sentences over their synthetic counterparts (Fig.~\ref{fig:fig3_synthetic_binarized}b), even when the synthetic sentences were formed such that the stronger models (i.e., GPT-2, RoBERTA, or ELECTRA) favored them over the natural sentences; see Extended Data Table~\ref{tab:ext_table_1_natural_vs_synthetic_sentence_pairs} for examples. Evaluating natural sentence preference separately for each model-pairing (Extended Data Fig.~\ref{fig:extfig_4_natural_vs_synthwise_pairwise_model_comparison}), we find that these imperfections can be uncovered even when pairing a strong model with a relatively weak model (such that the strong model ``accepts'' the synthetic sentence and the weak model rejects it).

\FloatBarrier

\begin{table}[t]
\centering
\scriptsize
 
\begin{adjustbox}{center}
\begin{tabularx}{18cm}{lllc}
\toprule
                                                          sentence &                               log probability (model 1) &                               log probability (model 2) &   \# human choices \\
\midrule
                   $s_1$: You can reach his stories on an instant. &             $\log p(s_1 | \textrm{GPT-2})=$\num{-64.92} &  $\log p(s_1 | \textrm{RoBERTa})=$\textbf{\num{-59.98}} &  \textbf{\num{10}} \\
               $s_2$: Anybody can behead a rattles an an antelope. &    $\log p(s_2 | \textrm{GPT-2})=$\textbf{\num{-40.45}} &           $\log p(s_2 | \textrm{RoBERTa})=$\num{-90.87} &            \num{0} \\\midrule
             $s_1$: However they will still compare you to others. &           $\log p(s_1 | \textrm{RoBERTa})=$\num{-53.40} &    $\log p(s_1 | \textrm{GPT-2})=$\textbf{\num{-31.59}} &  \textbf{\num{10}} \\
             $s_2$: Why people who only give themselves to others. &  $\log p(s_2 | \textrm{RoBERTa})=$\textbf{\num{-48.66}} &             $\log p(s_2 | \textrm{GPT-2})=$\num{-47.13} &            \num{0} \\\midrule
      $s_1$: He healed faster than any professional sports player. &           $\log p(s_1 | \textrm{ELECTRA})=$\num{-48.77} &     $\log p(s_1 | \textrm{BERT})=$\textbf{\num{-50.21}} &  \textbf{\num{10}} \\
                   $s_2$: One gets less than a single soccer team. &  $\log p(s_2 | \textrm{ELECTRA})=$\textbf{\num{-38.25}} &              $\log p(s_2 | \textrm{BERT})=$\num{-59.09} &            \num{0} \\\midrule
                   $s_1$: That is the narrative we have been sold. &              $\log p(s_1 | \textrm{BERT})=$\num{-56.14} &    $\log p(s_1 | \textrm{GPT-2})=$\textbf{\num{-26.31}} &  \textbf{\num{10}} \\
                      $s_2$: This is the week you have been dying. &     $\log p(s_2 | \textrm{BERT})=$\textbf{\num{-50.66}} &             $\log p(s_2 | \textrm{GPT-2})=$\num{-39.50} &            \num{0} \\\midrule
        $s_1$: The resilience is made stronger by early adversity. &               $\log p(s_1 | \textrm{XLM})=$\num{-62.95} &  $\log p(s_1 | \textrm{RoBERTa})=$\textbf{\num{-54.34}} &  \textbf{\num{10}} \\
                $s_2$: Every thing is made alive by infinite Ness. &      $\log p(s_2 | \textrm{XLM})=$\textbf{\num{-42.95}} &           $\log p(s_2 | \textrm{RoBERTa})=$\num{-75.72} &            \num{0} \\\midrule
        $s_1$: President Trump threatens to storm the White House. &              $\log p(s_1 | \textrm{LSTM})=$\num{-58.78} &  $\log p(s_1 | \textrm{RoBERTa})=$\textbf{\num{-41.67}} &  \textbf{\num{10}} \\
               $s_2$: West Surrey refused to form the White House. &     $\log p(s_2 | \textrm{LSTM})=$\textbf{\num{-40.35}} &           $\log p(s_2 | \textrm{RoBERTa})=$\num{-67.32} &            \num{0} \\\midrule
                 $s_1$: Las beans taste best with a mustard sauce. &              $\log p(s_1 | \textrm{RNN})=$\num{-131.62} &  $\log p(s_1 | \textrm{RoBERTa})=$\textbf{\num{-60.58}} &  \textbf{\num{10}} \\
          $s_2$: Roughly lanes being alive in a statement ratings. &      $\log p(s_2 | \textrm{RNN})=$\textbf{\num{-49.31}} &           $\log p(s_2 | \textrm{RoBERTa})=$\num{-99.90} &            \num{0} \\\midrule
           $s_1$: You are constantly seeing people play the multi. &           $\log p(s_1 | \textrm{3-gram})=$\num{-107.16} &  $\log p(s_1 | \textrm{ELECTRA})=$\textbf{\num{-44.79}} &  \textbf{\num{10}} \\
 $s_2$: This will probably the happiest contradicts the hypocrite. &   $\log p(s_2 | \textrm{3-gram})=$\textbf{\num{-91.59}} &           $\log p(s_2 | \textrm{ELECTRA})=$\num{-75.83} &            \num{0} \\\midrule
                    $s_1$: A buyer can own a genuine product also. &           $\log p(s_1 | \textrm{2-gram})=$\num{-127.35} &  $\log p(s_1 | \textrm{ELECTRA})=$\textbf{\num{-40.21}} &  \textbf{\num{10}} \\
       $s_2$: One versed in circumference of highschool I rambled. &  $\log p(s_2 | \textrm{2-gram})=$\textbf{\num{-113.73}} &           $\log p(s_2 | \textrm{ELECTRA})=$\num{-92.61} &            \num{0} \\
\bottomrule
\end{tabularx}
\end{adjustbox}
\scriptsize
\caption{\textbf{Examples of controversial synthetic-sentence pairs that maximally contributed to each model's prediction error.} For each model (double row, ``model 1''), the table shows results for two sentences on which the model failed severely. In each case, the failing model 1 prefers sentence $s_2$ (higher log probability bolded), while the model it was pitted against (``model 2'') and all 10 human subjects presented with that sentence pair prefer sentence $s_1$. (When more than one sentence pair induced an equal maximal error in a model, the example included in the table was chosen at random.)\label{tab:tab2_synthetic_controversial_sentence_pairs}}
\end{table}

\FloatBarrier

\subsection{Evaluating the entire dataset reveals a hierarchy of models}
\label{ssec:all_trial_results}
\begin{figure}[!t]
    \includegraphics{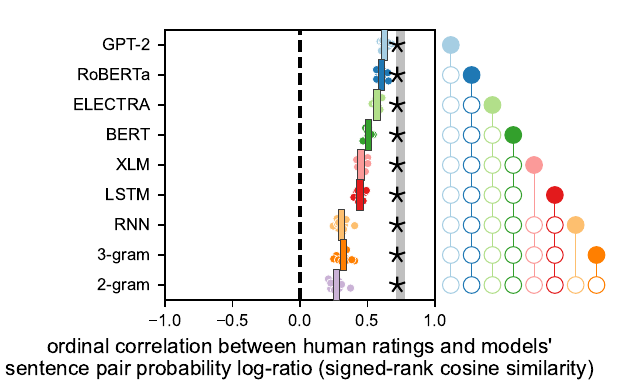}
	\caption{\textbf{Ordinal correlation of the models' sentence probability log-ratios and human Likert ratings.} For each sentence pair, model prediction was quantified by $\log \frac{p(s_1 \mid m)}{\ p(s_2 \mid m)}$. This log-ratio was correlated with the Likert ratings of each particular participant, using signed-rank cosine similarity (see 
	Methods). This analysis, taking all trials and human confidence level into account, indicates that GPT-2 performed best in predicting human sentence probability judgments. However, its predictions are still significantly misaligned with the human choices. See Fig.~\ref{fig:fig1_natural_binarized_accuracy}'s caption for details on the visualization convention.}
	\label{fig:fig4_all_data_accuracy}
\end{figure}

\begin{figure}[!ht]
\includegraphics[width=1\linewidth]{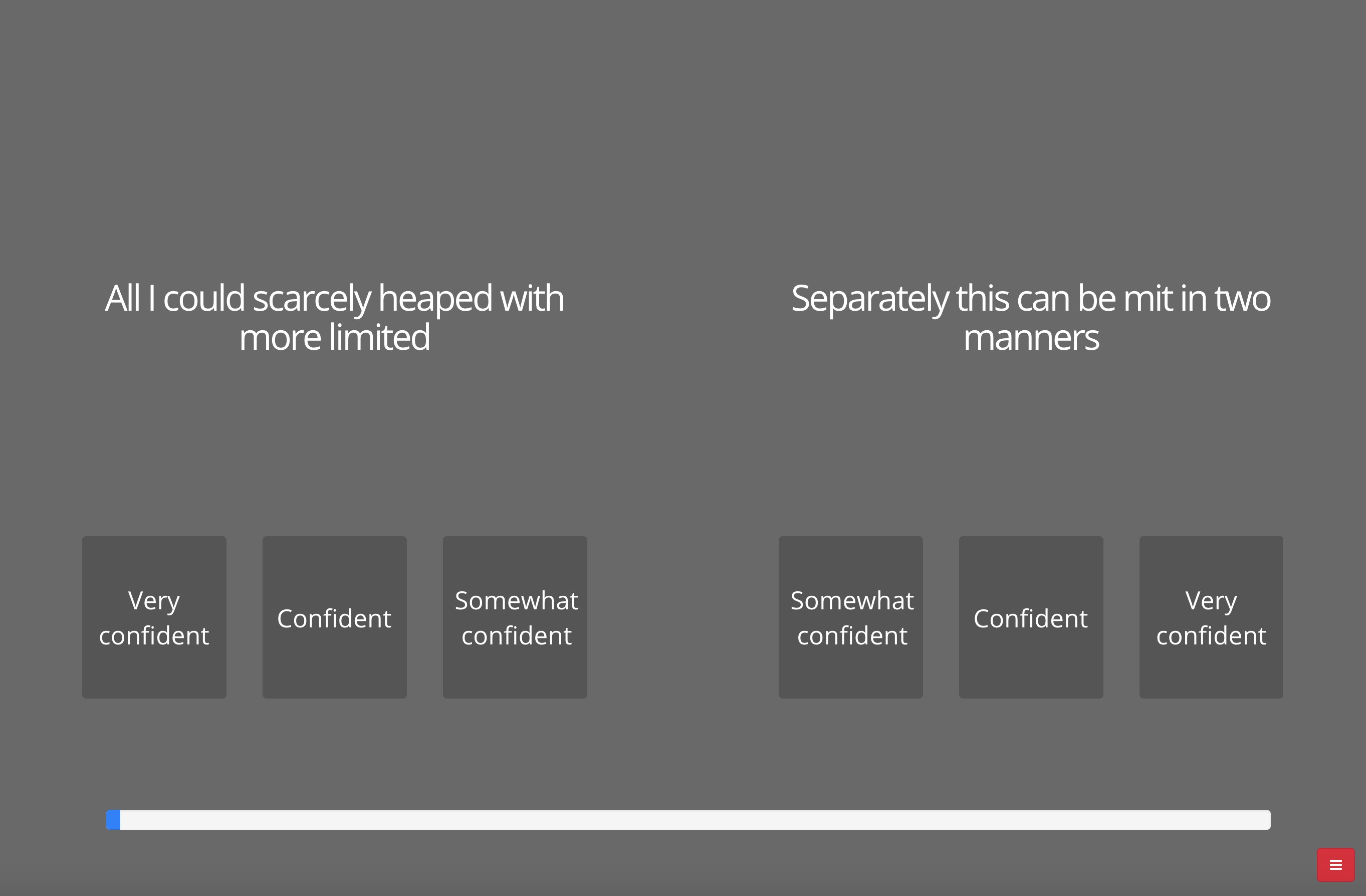}
\caption{\textbf{An example of one experimental trial, as presented to the participants}. The participant must choose one sentence while providing their confidence rating on a 3-point scale.}
\label{fig:extfig_1_example_trial}
\end{figure}

\begin{SCfigure}[1.2][!h]
    \includegraphics[]{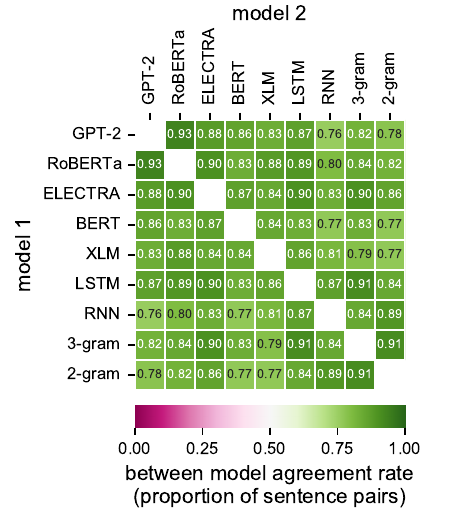}
	\caption{\textbf{Between-model agreement rate on the probability ranking of the 90 randomly sampled and paired natural sentence pairs evaluated in the experiment}. Each cell represents the proportion of sentence pairs for which two models make congruent probability ranking (i.e., both models assign a higher probability to sentence 1, or both models assign a higher probability to sentence 2).}
	\label{fig:extfig_2_randomly_sampled_sents_model_agreement}
\end{SCfigure}

\begin{figure}[!h]
\includegraphics[]{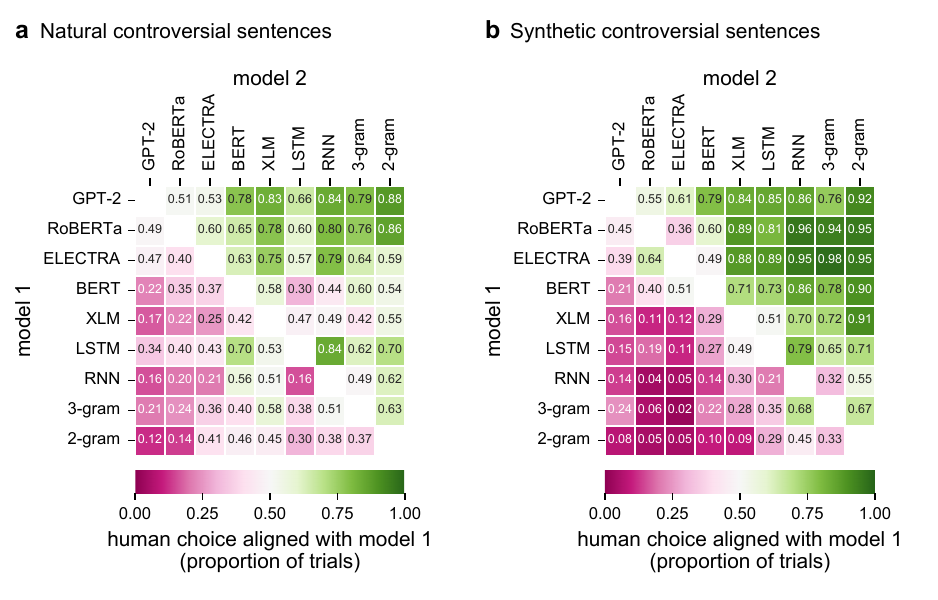}
 \caption{\textbf{Pairwise model comparison of model-human consistency.} For each pair of models (represented as one cell in the matrices above), the only trials considered were those in which the stimuli were either selected (a) or synthesized (b) to contrast the predictions of the two models. For these trials, the two models always made controversial predictions (i.e., one sentence is preferred by the first model and the other sentence is preferred by the second model). The matrices above depict the proportion of trials in which the binarized human judgments aligned with the row model (``model 1"). For example, GPT-2 (top-row) was always more aligned (green hues) with the human choices than its rival models. In contrast, 2-gram (bottom-row) was always less aligned (purple hues) with the human choices than its rival models.}
	\label{fig:extfig_3_pairwise_model}
\end{figure}

\begin{SCfigure}[1.4][!ht] %
    \includegraphics{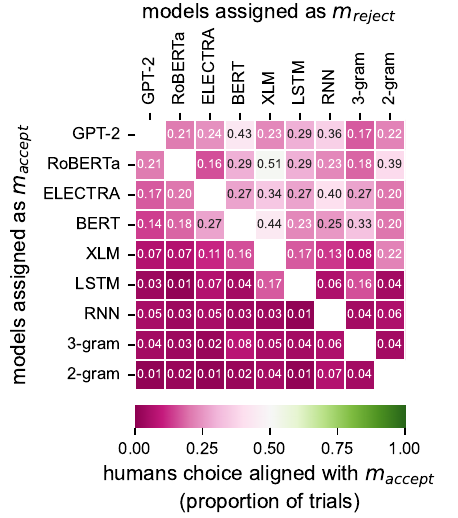} 
	\caption{\textbf{Pairwise model analysis of human response for natural vs. synthetic sentence pairs.} In each optimization condition, a synthetic sentence $s$ was formed by modifying a natural sentence $n$ so the synthetic sentence would be ``rejected" by one model ($m_{reject}$, columns), minimizing $p(s \mid m_{reject})$, and would be ``accepted" by another model ($m_{accept}$, rows), satisfying the constraint 
	$p(s \mid m_{accept})\geq p(n \mid m_{accept})$. Each cell above summarizes model-human agreement in trials resulting from one such optimization condition. The color of each cell denotes the proportion of trials in which humans judged a synthetic sentence to be more likely than its natural counterpart and hence aligned with $m_{accept}$. For example, the top-right cell depicts human judgments for sentence pairs formed to minimize the probability assigned to the synthetic sentence by the simple 2-gram model while ensuring that GPT-2 would judge the synthetic sentence to be at least as likely as the natural sentence; humans favored the synthetic sentence in only 22 out the 100 sentence pairs in this condition.}
	\label{fig:extfig_4_natural_vs_synthwise_pairwise_model_comparison}
\end{SCfigure}

Rather than evaluating each model's prediction accuracy with respect to the particular sentence pairs that were formed to compare this model to alternative models, we can maximize our statistical power by computing the average prediction accuracy for each model with respect to all of the experimental trials we collected. Furthermore, rather than binarizing the human and model judgments, here we measure the ordinal correspondence between the graded human choices (taking confidence into account) and the log ratio of the sentence probabilities assigned by each candidate model. Using this more sensitive benchmark (Fig.~\ref{fig:fig4_all_data_accuracy}), we found GPT-2 to be the most human-aligned, followed by RoBERTa; then ELECTRA; BERT; XLM and LSTM; and the RNN, 3-gram, and 2-gram models. However, all of the models (including GPT-2) were found to be significantly less accurate than the lower bound on the noise ceiling.

\begin{figure}[htbp]
	\centering
	\includegraphics{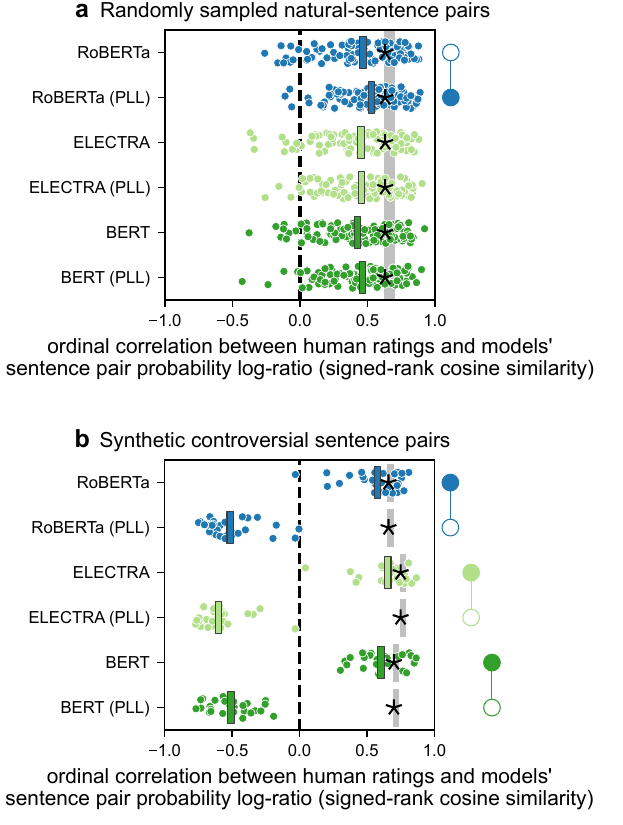}
	\caption{\textbf{Human consistency of bidirectional transformers: approximate log-likelihood versus pseudo-log-likelihood (PLL).} Each dot in the plots above depicts the ordinal correlation between the judgments of one participant and the predictions of one model. \textbf{(a)} The performance of BERT, RoBERTa, and ELECTRA in predicting the human judgments of randomly sampled natural sentence pairs in the main experiment, using two different likelihood measures: our novel approximate likelihood method (i.e., averaging multiple conditional probability chains, see Methods) and pseudo-likelihood (PLL, summating the probability of each word given all of the other words \cite{wang-cho-2019-bert}). For each model, we statistically compared the two likelihood measures to each other and to the noise ceiling using a two-sided Wilcoxon signed-rank test across the participants. False discovery rate was controlled at $q < 0.05$ for the 9 comparisons. \textbf{When predicting human preferences of natural sentences, the pseudo-log-likelihood measure is at least as accurate as our proposed approximate log-likelihood measure.} \textbf{(b)} Results from a follow-up experiment, in which we synthesized synthetic sentence pairs for each of the model pairs, pitting the two alternative likelihood measures against each other. Statistical testing was conducted in the same fashion as in panel a. These results indicate that for each of the three bidirectional language models, the approximate log-likelihood measure is considerably and significantly ($q < 0.05$) more human-consistent than the pseudo-likelihood measure. \textbf{Synthetic controversial sentence pairs uncover a dramatic failure mode of the pseudo-log-likelihood measure, which remains covert when the evaluation is limited to randomly-sampled natural sentences.} See Extended Data Table \ref{tab:ext_table_2_synthetic_controversial_sentence_pairs_PLL_followup} for synthetic sentence pair examples.}
    \label{fig:extfig_5_has_a_mouth_experiment}
\end{figure}

One possible reason for the poorer performance of the bidirectional transformers (RoBERTa, ELECTRA, BERT, and XLM) compared to the unidirectional transformer (GPT-2) is that computing sentence probabilities in these models is complex, and the probability estimator we developed (see Methods, ``Evaluating sentence probabilities in transformer models'') could be non-optimal; Indeed, the popular pseudo-log-likelihood (PLL) approach yields slightly higher accuracy for randomly sampled natural-sentence pairs (Extended Data Fig.~\ref{fig:extfig_5_has_a_mouth_experiment}a). And yet, when we directly compared our estimator to PLL by means of generating and administrating new synthetic controversial sentences, our estimator was found to be markedly better aligned to human judgments (Extended Data Fig.~\ref{fig:extfig_5_has_a_mouth_experiment}b and Extended Data Table~\ref{tab:ext_table_2_synthetic_controversial_sentence_pairs_PLL_followup}).

Finally, a control analysis employing probability measures normalized by token count revealed that such normalization had minimal influence on the observed differences among models (Supplementary Section \ref{sup_results_normalization} and Supplementary Fig.~\ref{fig:s1_token_count_normalized_prediction_accuracy}).

\
\section{Discussion}\label{sec12}

In this study, we probed the ability of language models to predict human relative sentence probability judgments using controversial sentence pairs, selected or synthesized so that two models disagreed about which sentence was more probable. We found that (1) GPT-2 (a unidirectional transformer model trained on predicting upcoming tokens) and RoBERTa (a bidirectional transformer trained on a held-out token prediction task) were the most predictive of human judgments on controversial natural-sentence pairs (Fig.~\ref{fig:fig1_natural_binarized_accuracy}b); (2) GPT-2, RoBERTa, and ELECTRA (a bidirectional transformer trained on detecting corrupted tokens) were the most predictive of human judgments on pairs of sentences synthesized to maximize controversiality (Fig.~\ref{fig:fig3_synthetic_binarized}a); and (3) GPT-2 was the most human-consistent model when considering the entire behavioral dataset we collected (Fig.~\ref{fig:fig4_all_data_accuracy}). And yet, all of the models, including GPT-2, exhibited behavior inconsistent with human judgments; using an alternative model as a counterforce, we could corrupt natural sentences such that their probability under a model did not decrease, but humans tended to reject the corrupted sentence as unlikely (Fig.~\ref{fig:fig3_synthetic_binarized}b).

\subsection{Implications for computational psycholinguistic modeling}

Unlike convolutional neural networks, whose architectural design principles are roughly inspired by biological vision \cite{lindsay2021convolutional}, the design of current neural network language models is largely uninformed by psycholinguistics and neuroscience. And yet, there is an ongoing effort to adopt and adapt neural network language models to serve as computational hypotheses of how humans process language, making use of a variety of different architectures, training corpora, and training tasks \cite{wehbe2014aligning,Mariya2019Interpreting,heilbron_hierarchy_2022,jain2020interpretable,Lyu2021finding,schrimpf2021neural,wilcox-etal-2021-targeted,goldstein_shared_2022,caucheteux_brains_2022,Arehalli22}. We found that recurrent neural networks make markedly human-inconsistent predictions once pitted against transformer-based neural networks. This finding coincides with recent evidence that transformers also outperform recurrent networks for predicting neural responses as measured by ECoG or fMRI \cite{schrimpf2021neural,goldstein_shared_2022}, as well as with evidence from model-based prediction of human reading speed \cite{wilcox-etal-2021-targeted,merkx2021human} and N400 amplitude \cite{merkx2021human,Michaelov2021Different}. Among the transformers, GPT-2, RoBERTa, and ELECTRA showed the best performance. These models are trained to optimize only word-level prediction tasks, as opposed to BERT and XLM which are additionally trained on next-sentence prediction and cross-lingual tasks, respectively (and have the same architecture as RoBERTa). This suggests that local word prediction provides better alignment with human language comprehension.

Despite the agreement between our results and previous work in terms of model ranking, the significant failure of GPT-2 in predicting the human responses to natural versus synthetic controversial pairs (Fig.~\ref{fig:fig3_synthetic_binarized}b) demonstrates that GPT-2 does not fully emulate the computations employed in human processing of even short sentences. This outcome is in some ways unsurprising, given that GPT-2 (like all of the other models we considered) is an off-the-shelf machine learning model that was not designed with human psycholinguistic and physiological details in mind. And yet, the considerable human inconsistency we observed seems to stand in stark contrast with the recent report of GPT-2 explaining about 100 percent of the explainable variance in fMRI and ECoG responses to natural sentences \cite{schrimpf2021neural}. Part of this discrepancy could be explained by the fact that Schrimpf and colleagues \cite{schrimpf2021neural} mapped GPT-2 hidden-layer activations to brain data by means of regularized linear regression, which can identify a subspace within GPT-2's language representation that is well-aligned with brain responses even if GPT-2's overall sentence probabilities are not human-like. More importantly, when language models are evaluated with natural language, strong statistical models might capitalize on features in the data that are distinct from, but highly correlated with, features that are meaningful to humans. Therefore, a model that performs well on typical sentences might employ computational mechanisms that are very distinct from the brain's, which will only be revealed by testing the model in a more challenging domain. Note that even the simplest model we considered---a 2-gram frequency table---actually performed quite well on predicting human judgments for randomly-sampled natural sentences, and its deficiencies only became obvious when challenged by controversial sentence pairs. We predict that there will be substantial discrepancies between neural representations and current language models when using stimuli that intentionally stress-test this relationship, using our proposed sentence-level controversiality approach or complementary ideas such as maximizing controversial transition probabilities between consecutive words \cite{rakocevic2021synthesizing}.

Using controversial sentences can be seen as a generalization test of language models: can models predict what kinds of changes to a natural sentence will lead to humans rejecting the sentence as improbable? Humans are sometimes capable of comprehending language with atypical constructions (e.g. in cases when pragmatic judgments can be made about a speaker's intentions from environmental and linguistic context \cite{goodman2016pragmatic}), but none of the models we tested were fully able to predict which syntactic or semantic perturbations would be accepted or rejected by humans. One possibility is that stronger next-word prediction models, using different architectures, learning rules, or training data, might close the gap between models and humans. Alternatively, it might be that optimizing for other linguistic tasks, or even non-linguistic task demands (in particular, representing the external world, the self, and other agents) will turn out to be critical for achieving human-like natural language processing \cite{howell2005model}.

\subsection{Controversial sentence pairs as adversarial attacks}

Machine vision models are highly susceptible to adversarial examples \cite{szegedy_intriguing_2013, goodfellow_explaining_2015}. Such adversarial examples are typically generated by choosing a correctly classified natural image and then searching for a minuscule (and therefore human-imperceptible) image perturbation that would change the targeted model's classification. The prospect that similar covert model failure modes may exist also for language models has motivated proposed generalizations of adversarial methods to textual inputs \cite{zhang2020adversarial}. However, imperceptible perturbations cannot be applied to written text: any modified word or character is humanly perceptible. Prior work on adversarial examples for language models have instead relied on heuristic constraints aiming to limit the change in the meaning of the text, such as flipping a character \cite{Liang2017deep,ebrahimi2018hotflip}, changing number or gender \cite{abdou2020sensitivity}, or replacing words with synonyms \cite{alzantot-etal-2018-generating,ribeiro2018semantically,ren-etal-2019-generating}. However, since these heuristics are only rough approximations of human language processing, many of these methods fail to preserve semantic meaning \cite{morris2020reevaluating}. Interactive (``human-in-the-loop'') adversarial approaches allow human subjects to repeatedly alter model inputs such that it confuses target models but not secondary participants \cite{wallace2019trick,kiela2021dynabench}, but these approaches are inherently slow and costly and are limited by mental models the human subjects form about the evaluated language models.

By contrast, testing language models on controversial sentence pairs does not require approximating or querying a human ground truth during optimization---the objective of controversiality is independent of correctness. Instead, by designing inputs to elicit conflicting predictions among the models and assessing human responses to these inputs only once the optimization loop has terminated, we capitalize on the simple fact that if two models disagree with respect to an input, at least one of the models must be making an incorrect prediction. Pitting language models against other language models also can be conducted by other approaches such as ``red-teaming'', where an alternative language model is used as a generator of potential adversarial examples for a targeted model and a classifier is used to filter the generated examples such that the output they induce in the targeted model is indeed incorrect \cite{perez2022red}. Our approach shares the underlying principle that an alternative language model can drive a more powerful test than handcrafted heuristics, but here the models have symmetric roles (there are no ``attacking'' and ``attacked'' models) and we can optimize stimuli directly without relying on filtering.

\subsection{Limitations and future directions}

While our results demonstrate that using controversial stimuli can identify subtle differences in language models' alignment with human judgments, our study was limited in a number of ways. Our stimuli were all 8-word English sentences, limiting our ability to make cognitively meaningful claims that apply to language use globally. 8-word sentences are long enough to include common syntactic constructions and convey meaningful ideas but may not effectively probe long-distance syntactic dependencies \cite{GIBSON19981}. Future work may introduce additional sentence lengths and languages, as well as (potentially adaptive) controversial sentence optimization procedures that consider large sets of candidate models, allowing for greater model coverage than our simpler pairwise approach. Future work may also complement the model-comparative experimental design with procedures designed to identify potential failure modes common to all models.

A more substantial limitation of the current study is that, like any comparison of pre-trained neural networks as potential models of human cognition, there could be multiple reasons (i.e., training data, architecture, training tasks, learning rules) why particular models are better aligned with human judgments. For example, as we did not systematically control the training corpora used for training the models, it is possible that some of the observed differences are due to differences in the training sets rather than model architecture. Therefore, while our results expose failed model predictions, they do not readily answer why these failed predictions arise. Future experiments could compare custom-trained or systematically manipulated models, which reflect specific hypotheses about human language processing. In Extended Data Fig.~\ref{fig:extfig_5_has_a_mouth_experiment}, we demonstrate the power of using synthetic controversial stimuli to conduct sensitive comparisons between models with subtle differences in how sentence probabilities are calculated.

It is important to note that our analyses considered human relative probability judgments as reflecting a scalar measure of acceptability. We made this assumption in order to bring the language models (which assign a probability measure to each sentence) and the human participants onto a common footing. However, it is possible that different types of sentence pairs engage different human cognitive processes. For pairs of synthetic sentences, both sentences may be unacceptable in different ways (e.g. exhibit different kinds of grammatical violations), requiring a judgment that weighs the relative importance of multiple dimensions \cite{Watt1975-rq} and could therefore produce inconsistent rankings across participants or across trials \cite{schutze2016empirical}. 
By contrast, asking participants to compare a natural and a synthetic sentence (Fig.~\ref{fig:fig3_synthetic_binarized}b, Extended Data Table~\ref{tab:ext_table_1_natural_vs_synthetic_sentence_pairs}) may be more analogous to previous work measuring human acceptability judgments for sentence pairs \cite{schütze_sprouse_2014}. Nonetheless, it is worth noting that for \emph{all} of the controversial conditions, the noise ceiling was significantly above the models' prediction accuracy, indicating non-random human preferences unexplained by current models that should be accounted for by future models, which may have to be more complex and capture multiple processes.

Finally, the use of synthetic controversial sentences can be extended beyond probability judgments. A sufficiently strong language model may enable constraining the experimental design search-space to particular sentence distributions (e.g., movie reviews or medical questions). Given such a constrained space, we may be able to search for well-formed sentences that elicit contradictory predictions in alternative domain-specific models (e.g., sentiment classifiers or question-answering models). However, as indicated by our results, the task of capturing distributions of well-formed sentences is less trivial than it seems.

\section{Methods}\label{sec11}

\subsection{Language models}

We tested nine models from three distinct classes: n-gram models, recurrent neural networks, and transformers. The n-gram models were trained with open source code from the Natural Language Toolkit \cite{bird2009natural}, the recurrent neural networks were trained with architectures and optimization procedures available in PyTorch \cite{NEURIPS2019_9015}, and the transformers were implemented with the open-source repository HuggingFace \cite{wolf2020transformers}. For full details see Supplementary Section \ref{sup_methods_language_models}.

\subsection{Evaluating sentence probabilities in transformer models}

We then sought to compute the probability of arbitrary sentences under each of the models described above. The term ``sentence'' is used in this context in its broadest sense---a sequence of English words, not necessarily restricted to grammatical English sentences. Unlike some classification tasks in which valid model predictions may be expected only for grammatical sentences (e.g., sentiment analysis), the sentence probability comparison task is defined over the entire domain of eight-word sequences.

For the set of unidirectional models, evaluating sentence probabilities was performed simply by summing the log probabilities of each successive token in the sentence from left to right, given all the previous tokens. For bidirectional models, this process was not as straightforward. One challenge is that transformer model probabilities do not necessarily reflect a coherent joint probability; the summed log sentence probability resulting from adding words in one order (e.g. left to right) does not necessarily equal the probability resulting from a different order (e.g. right to left). Here we developed a novel formulation of bidirectional sentence probabilities in which we considered all permutations of serial word positions as possible construction orders (analogous to the random word visitation order used to sample serial reproduction chains, \cite{Yamakoshi2022}). In practice, we observed that the distribution of log probabilities resulting from different permutations tends to center tightly around a mean value (for example, for RoBERTa evaluated with natural sentences, the average coefficient of variation was approximately $0.059$). Therefore in order to efficiently calculate bidirectional sentence probability, we evaluate 100 different random permutations and define the overall sentence log probability as the mean log probability calculated from each permutation. Specifically, we initialized an eight-word sentence with all tokens replaced with the ``mask'' token used in place of to-be-predicted words during model training. We selected a random permutation $P$ of positions 1 through 8, and started by computing the probability of the word at first of these positions $P_1$ given the other seven ``mask'' tokens. We then replaced the ``mask'' at position $P_1$ with the actual word at this position and computed the probability of the word at $P_2$ given the other six ``mask'' tokens and the word at $P_1$. This process was repeated until all ``mask'' tokens had been filled by the corresponding word. 

A secondary challenge in evaluating sentence probabilities in bidirectional transformer models stems from the fact that these models use word-piece tokenizers (as opposed to whole words), and that these tokenizers are different for different models. For example, one tokenizer might include the word ``beehive'' as a single token, while others strive for a smaller library of unique tokens by evaluating ``beehive'' as the two tokens ``bee'' and ``hive''. The model probability of a multi-token word---similar to the probability of a multi-word sentence---may depend on the order in which the chain rule is applied. Therefore, all unique permutations of token order for each multi-token word were also evaluated within their respective ``masks''. For example, the probability of the word ``beehive'' would be evaluated as follows:

\begin{equation}
\begin{split}
\log p(w=\textrm{beehive}) = & 0.5 \big(\log p(w_1 = \textrm{bee} \mid w_2 = \textrm{MASK})+\log p(w_2 = \textrm{hive} \mid w_1=\textrm{bee})\big)\\ + & 0.5 \big(\log p(w_2 = \textrm{hive} \mid w_1=\textrm{MASK})+\log p(w_1 = \textrm{bee} \mid w_2 = \textrm{hive})\big)
\end{split}
\end{equation}

This procedure aimed to yield a more fair estimate of the conditional probabilities of word-piece tokens and therefore the overall probabilities of multi-token words by 1) ensuring that the word-piece tokens were evaluated within the same context of surrounding words and masks, and 2) eliminating the bias of evaluating the word-piece tokens in any one particular order in models which were trained to predict bidirectionally. 

One more procedure was applied in order to ensure that all models were computing a probability distribution over sentences with exactly 8 words. When evaluating the conditional probability of a masked word in models with word-piece tokenizers, we normalized the model probabilities to ensure that only single words were being considered, rather than splitting the masked tokens into multiple words. At each evaluation step, each token was restricted to come from one of four normalized distributions: i) single-mask words were restricted to be tokens with appended white space, ii) masks at the beginning of a word were restricted to be tokens with preceding white space (in models with preceding white space, e.g. BERT), iii) masks at the end of words were restricted to be tokens with trailing white space (in models with trailing white space, e.g. XLM), and iv) masks in the middle of words were restricted to tokens with no appended white space.

\subsection{Assessing potential token count effects on sentence probabilities}
Note that, because tokenization schemes varied across models, the number of tokens in a sentence could differ for different models. These alternative tokenizations can be conceived of as different factorizations of the modeled language distribution, changing how a sentence's log probability is additively partitioned across the conditional probability chain (but not affecting its overall probability) \cite{chestnut_2019_perplexity}. Had we attempted to normalize across models by dividing the log probability by the number of tokens, as is often done when aligning model predictions to human acceptability ratings \cite{lau_2017_grammaticality,lau_2020_how}, our probabilities would have become strongly tokenization-dependent \cite{chestnut_2019_perplexity}. To empirically confirm that tokenization differences were not driving our results, we statistically compared the token counts of each model's preferred synthetic sentences with the token counts of their non-preferred counterparts. While we found significant differences for some of the models, there was no systematic association between token count and model sentence preferences (Supplementary Table~\ref{tab:s1_token_count_analysis}). In particular, lower sentence probabilities were not systematically confounded by higher token counts.

\subsection{Defining a shared vocabulary}
To facilitate the sampling, selection, and synthesis of sentences that could be evaluated by all of the candidate models, we defined a shared vocabulary of \num{29157} unique words. Defining this vocabulary was necessary in order to unify the space of possible sentences between the transformer models (which can evaluate any input due to their word-piece tokenizers) and the neural network and n-gram models (which include whole words as tokens), and to ensure we only included words that were sufficiently prevalent in the training corpora for all models. The vocabulary consisted of the words in the subtlex database \cite{van2014subtlex}, after removing words that occurred fewer than 300 times in the 300M word corpus (see Supplementary Section \ref{sup_methods_language_models}) used to train the n-gram and recurrent neural network models (i.e., with frequencies lower than one in a million).

\subsection{Sampling of natural sentences}

Natural sentences were sampled from the same four text sources used to construct the training corpus for the n-gram and recurrent neural network models, while ensuring that there was no overlap between training and testing sentences. Sentences were filtered to include only those with eight distinct words and no punctuation aside from periods, exclamation points, or question marks at the end of a sentence. Then, all eight-word sentences were further filtered to include only the words included in the shared vocabulary and to exclude those included in a predetermined list of inappropriate words and phrases. To identify controversial pairs of natural sentences, we used integer linear programming to search for sentences that had above-median probability in one model and minimum probability rank in another model (see Supplementary Section \ref{sup_methods_selection}).

\subsection{Generating synthetic controversial sentence pairs} \label{ssec:generating_synthetic}
For each pair of models, we synthesized 100 sentence triplets. Each triplet was initialized with a natural sentence $n$ (sampled from Reddit). The words in sentence $n$ were iteratively modified to generate a synthetic sentence with reduced probability according to the first model but not according to the second model. This process was repeated to generate another synthetic sentence from $n$, in which the roles of the two models were reversed. Conceptually, this approach resembles Maximum Differentiation (MAD) competition \cite{wang_maximum_2008}, introduced to compare models of image quality assessment. Each synthetic sentence was generated as a solution for a constrained minimization problem:

\begin{equation}
\begin{aligned}
\label{eq:s1_loss}
    s^* = \argmin_s & \log p(s \mid m_{reject})\\
    \textrm{subject to} & \log p(s \mid m_{accept}) \geq \log p(n \mid m_{accept})
\end{aligned}
\end{equation}
$m_{reject}$ denotes the model targeted to assign reduced sentence probability to the synthetic sentence compared to the natural sentence, and $m_{accept}$ denotes the model targeted to maintain a synthetic sentence probability greater or equal to that of the natural sentence. For one synthetic sentence, one model served as $m_{accept}$ and the other model served as $m_{reject}$, and for the other synthetic sentence the model roles were flipped.

At each optimization iteration, we selected one of the eight words pseudorandomly (so that all eight positions would be sampled $N$ times before any position would be sampled $N+1$ times) and searched the shared vocabulary for the replacement word that would minimize the $\log p(s \mid m_{reject})$ under the constraint. We excluded potential replacement words that already appeared in the sentence, except for a list of 42 determiners and prepositions such as ``the", ``a", or ``with", which were allowed to repeat. The sentence optimization procedure was concluded once eight replacement attempts (i.e., words for which no loss-reducing replacement has been found) have failed in a row.

\subsection{Word-level search for bidirectional models}
For models for which the evaluation of $\log p(s \mid m)$ is computationally cheap (2-gram, 3-gram, LSTM, and the RNN), we directly evaluated the log-probability of the \num{29157} sentences resulting from each of the \num{29157} possible word replacements. When such probability vectors were available for both models, we simply chose the replacement minimizing the loss. For GPT-2, whose evaluation is slower, we evaluated sentence probabilities only for word replacements for which the new word had a conditional log-probability (given the previous words in the sentence) of no less than \num{-10}; in rare cases when this threshold yielded fewer than 10 candidate words, we reduced the threshold in steps of \num{5} until there were at least 10 words above the threshold. For the bi-directional models (BERT, RoBERTa, XLM, and ELECTRA), for which the evaluation of $\log p(s \mid m)$ is costly even for a single sentence, we used a heuristic to prioritize which replacements to evaluate.

Since bi-directional models are trained as masked language models, they readily provide word-level completion probabilities. These word-level log-probabilities typically have positive but imperfect correlation with the log-probabilities of the sentences resulting from each potential completion. We hence formed a simple linear regression-based estimate of $\log p(s\{i\} \gets w \mid m)$, the log-probability of the sentence $s$ with word $w$ assigned at position $i$, predicting it from $\log p(s\{i\}=w \mid m, s\{i\} \gets mask)$, the completion log-probability of word $w$ at position $i$, given the sentence with the i-th word masked:

\begin{equation}
\label{eq:word_level_regression}
\log \hat{p}(s\{i\}\gets w \mid m) = \beta_1 \log p(s\{i\}=w \mid m, s\{i\} \gets mask) + \beta_0
\end{equation}

This regression model was estimated from scratch for each word-level search. When a word was first selected for being replaced, the log-probability of two sentences was evaluated: the sentence resulting from substituting the existing word with the word with the highest completion probability and the sentence resulting from substituting the existing word with the word with the lowest completion probability. These two word-sentence log-probability pairs, as well as the word-sentence log-probability pair pertaining to the current word, were used to fit the regression line. The regression prediction, together with the sentence probability for the other model (either the exact probability, or approximate probability if the other model was also bi-directional) was used to predict $\log p(s \mid m_{reject})$ for each of the \num{29157} potential replacements. We then evaluated the true (non-approximate) sentence probabilities of the replacement word with the minimal predicted probability. If this word indeed reduced the sentence probability, it was chosen to serve as the replacement and the word-level search was terminated (i.e., proceeding to search a replacement for another word in the sentence). If it did not reduce the probability, the regression model (Eq.~\ref{eq:word_level_regression}) was updated with the new observation, and the next replacement expected to minimize the sentence probability was evaluated. This word-level search was terminated after five sentence evaluations that did not reduce the loss. 

\subsection{Selecting the best triplets from the optimized sentences}
Since the discrete hill-climbing procedure described above is highly local, the degree to which this succeeded in producing highly-controversial pairs varied depending on the starting sentence $n$. We found that typically, natural sentences with lower than average log-probability gave rise to synthetic sentences with greater controversiality. To better represent the distribution of natural sentences while still choosing the best (most controversial) triplets for human testing, we used stratified selection.

First, we quantified the controversiality of each triplet as 
\begin{equation}
\label{eq:triplet_controversiality}
c_{m_1,m_2}(n,s_1,s_2)=\log \frac{p(n \mid m_1)}{p(s_1 \mid m_1)}+ \log \frac{p(n \mid m_2)}{p(s_2 \mid m_2)},
\end{equation}
where $s_1$ is the sentence generated to reduce the probability in model $m_1$ and $s_2$ is the sentence generated to reduce the probability in model $m_2$.

We employed integer programming to choose the 10 most controversial triplets from the 100 triplets optimized for each model pair (maximizing the total controversiality across the selected triplets), while ensuring that for each model, there was exactly one natural sentence in each decile of the natural sentences probability distribution. The selected 10 synthetic triplets were then used to form 30 unique experimental trials per model pair, comparing the natural sentence with one synthetic sentence, comparing the natural sentence with the other synthetic sentence, and comparing the two synthetic sentences.

\subsection{Design of the human experiment}
Our experimental procedures were approved by the Columbia University Institutional Review Board (protocol number IRB-AAAS0252) and were performed in accordance with the approved protocol. All participants provided informed consent prior. We presented the controversial sentence pairs selected and synthesized by the language models to 100 native English-speaking, US-based participants (55 male) recruited from Prolific (\url{www.prolific.co}), and paid each participant \$5.95. The average participant age was 34.08 ± 12.32. The subjects were divided into 10 groups, and each ten-subject group was presented with a unique set of stimuli. Each stimulus set contained exactly one sentence pair from every possible combination of model pairs and the four main experimental conditions: selected controversial sentence pairs; natural vs. synthetic pair, where one model served as $m_{accept}$ and the other as $m_{reject}$; a natural vs. synthetic pair with the reverse model role assignments; and directly pairing the two synthetic sentences. These model-pair-condition combinations accounted for 144 (36$\times$4) trials of the task. In addition to these trials, each stimulus set also included nine trials consisting of sentence pairs randomly sampled from the database of eight-word sentences (and not already included in any of the other conditions). All subjects also viewed 12 control trials consisting of a randomly selected natural sentence and the same natural sentence with the words scrambled in a random order. The order of trials within each stimulus set as well as the left-right screen position of sentences in each sentence pair were randomized for all participants. While each sentence triplet produced by the optimization procedure (see subsection ``\nameref{ssec:generating_synthetic}'') gave rise to three trials, these were allocated such that no subject viewed the same sentence twice.

On each trial of the task, participants were asked to make a binary decision about which of the two sentences they considered more probable (for the full set of instructions given to participants, see Supplementary Fig.~\ref{fig:s2_taskinstructions}). In addition, they were asked to indicate one of three levels of confidence in their decision: somewhat confident, confident, or very confident. The trials were not timed, but a 90-minute time limit was enforced for the whole experiment. A progress bar at the bottom of the screen indicated to participants how many trials they had completed and had remaining to complete. 

 We rejected the data of 21 participants who failed to choose the original, unshuffled sentence in at least 11 of the 12 control trials, and acquired data from 21 alternative participants instead, all of whom passed this data-quality threshold. In general, we observed high agreement in sentence preferences among our participants, though the level of agreement varied across conditions. There was complete or near-complete agreement (at least 9/10 participants with the same binary sentence preference) in 52.2\% of trials for randomly-sampled natural-sentence pairs, 36.6\% of trials for controversial natural-sentence pairs, 67.6\% of trials for natural-synthetic pairs, and 60.0\% of trials for synthetic-synthetic pairs (versus a chance rate of 1.1\%, assuming a binomial distribution with $p=0.5$).

\subsection{Evaluation of model-human consistency}
\label{binarized_eval}
To measure the alignment on each trial between model judgments and human judgments, we binarized both measures; we determined which of the two sentences was assigned with a higher probability by the model, regardless of the magnitude of the probability difference, and which of the two sentences was favored by the subject, regardless of the reported confidence level. When both the subject and the model chose the same sentence, the trial was considered as correctly predicted by that model. This correctness measure was averaged across sentence pairs and across the 10 participants who viewed the same set of trials. For the lower bound on the noise ceiling, we predicted each subject's choices from a majority vote of the nine other subjects who were presented with the same trials. For the upper bound (i.e., the highest possible accuracy attainable on this data sample), we included the subject themselves in this majority vote-based prediction.

Since each of the 10 participant groups viewed a unique trial set, these groups provided 10 independent replications of the experiment. Models were compared to each other and to the lower bound of the noise ceiling by a Wilcoxon signed-rank test using these 10 independent accuracy outcomes as paired samples. For each analysis, the false discovery rate across multiple comparisons was controlled by the Benjamini-Hochberg procedure \cite{Benjamini1995Controlling}.

In Fig.~\ref{fig:fig4_all_data_accuracy}, we instead measure model-human consistency in a more continuous way, comparing the sentence probability ratio in a model to the graded Likert ratings provided by humans; see Supplementary Section \ref{sup_methods_ordinal} for full details.

\subsection{Selecting trials for model evaluation}
All of the randomly sampled natural-sentence pairs (Fig.~\ref{fig:fig1_natural_binarized_accuracy}a) were evaluated for each of the candidate models. Controversial sentence pairs (either natural, Fig.~\ref{fig:fig1_natural_binarized_accuracy}b or synthetic, Fig.~\ref{fig:fig3_synthetic_binarized}) were included in a model's evaluation set only if they were formed to target that model specifically. The overall summary analysis (Fig.~\ref{fig:fig4_all_data_accuracy}) evaluated all models on all available sentence pairs.

\subsection{Comparison to pseudo-log-likelihood acceptability measures}
\citet{wang-cho-2019-bert} proposed an alternative approach for computing sentence probabilities in bidirectional (BERT-like) models, using a pseudo-log-likelihood measure that simply sums the log-probability of each token conditioned on all of the other tokens in the sentence. While this measure does not reflect a true probability distribution \cite[][]{cho_2019_bert}, it is positively correlated with human acceptability judgments for several bidirectional models \cite{lau_2020_how,salazar-etal-2020-masked}. To directly compare this existing approach to our novel method for computing probabilities, we again used the method of controversial sentence pairs to identify the approach most aligned with human judgments. For each bidirectional model (BERT, RoBERTa, and ELECTRA), we created two copies of the model, each using a different approach for computing sentence probabilities. We synthesized 40 sentence pairs to maximally differentiate between the two copies of each model, with each copy assigning a higher probability to a different sentence in the pair. Subsequently, we tested 30 human participants, presenting each participant with all 120 sentence pairs. Model-human consistency was quantified as in the main experiment.

\subsection{Data and code availability}
The experimental stimuli, detailed behavioral testing results, sentence optimization code, and code for reproducing all analyses and figures are available at \href{https://github.com/dpmlab/contstimlang}{github.com/dpmlab/contstimlang} \cite{golan_siegelman_2023_code}.

\subsection*{Acknowledgments}
This material is based upon work partially supported by the National Science Foundation under Grant No. 1948004 to NK. This publication was made possible with the support of the Charles H. Revson Foundation to TG. The statements made and views expressed, however, are solely the responsibility of the authors. 

\subsection*{Author Contributions}
T.G., M.S., N.K., and C.B. designed the study. M.S. implemented the computational models and T.G. implemented the sentence pair optimization procedures. M.S. conducted the behavioral experiments. T.G. and M.S. analyzed the experiments' results. T.G., M.S., N.K., and C.B. wrote the paper.

\subsection*{Competing Interests}
The authors declare no competing interests.

\FloatBarrier
\printbibliography
\beginextdata

\FloatBarrier
\section{Extended Data}
\newgeometry{left=0.5in, top=0.5in, right=0.5in, bottom=1in}
\renewcommand{\figurename}{Extended Data Figure}
\renewcommand{\tablename}{Extended Data Table}

\begin{SCfigure}[1.4][ht]
	\includegraphics[trim={1cm 0 1cm 0},clip]{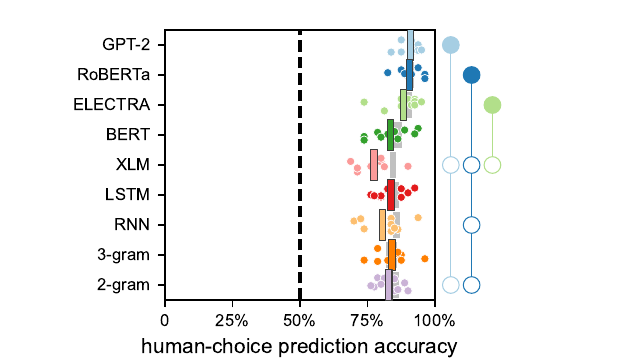}
 	\caption{\textbf{Model prediction accuracy for pairs of natural and synthetic sentences, evaluating each model across all of the sentence pairs in which it was targeted to rate the synthetic sentence to be less probable than the natural sentence.} The data binning applied here is complementary to the one used in Fig.~\ref{fig:fig3_synthetic_binarized}b, where each model was evaluated across all of the sentence pairs in which it was targeted to rate the synthetic sentence to be \emph{at least as probable} as the natural sentence. Unlike Fig.~\ref{fig:fig3_synthetic_binarized}b, where all of the models performed poorly, here no models were found to be significantly below the lower bound on the noise ceiling; typically, when a sentence was optimized to decrease its probability under any model (despite the sentence probability not decreasing under a second model), humans agreed that the sentence became less probable.}
 	\label{fig:extfig_6_synthetic_binarized_alternative_trial_selection}
\end{SCfigure}

\FloatBarrier

\begin{table}[ht]
\centering
\scriptsize
 
\begin{adjustbox}{center}
\begin{tabularx}{18cm}{lllc}
\toprule
                                                               sentence &                             log probability (model 1) &                             log probability (model 2) &   \# human choices \\
\midrule
                          $n$: I always cover for him and make excuses. &             $\log p(n | \textrm{GPT-2})=$\num{-36.46} &  $\log p(n | \textrm{2-gram})=$\textbf{\num{-106.95}} &  \textbf{\num{10}} \\
                            \makebox[0pt][l]{$s$: }\hphantom{$n$: }We either wish for it or ourselves do. &    $\log p(\mathmakebox[0pt][l]{s}\hphantom{n} | \textrm{GPT-2})=$\textbf{\num{-36.15}} &           $\log p(\mathmakebox[0pt][l]{s}\hphantom{n} | \textrm{2-gram})=$\num{-122.28} &            \num{0} \\\midrule
                         $n$: This is why I will never understand boys. &           $\log p(n | \textrm{RoBERTa})=$\num{-46.88} &  $\log p(n | \textrm{2-gram})=$\textbf{\num{-103.11}} &  \textbf{\num{10}} \\
                               \makebox[0pt][l]{$s$: }\hphantom{$n$: }This is why I will never kiss boys. &  $\log p(\mathmakebox[0pt][l]{s}\hphantom{n} | \textrm{RoBERTa})=$\textbf{\num{-46.75}} &           $\log p(\mathmakebox[0pt][l]{s}\hphantom{n} | \textrm{2-gram})=$\num{-107.91} &            \num{0} \\\midrule
                                $n$: One of the ones I did required it. &           $\log p(n | \textrm{ELECTRA})=$\num{-35.97} &     $\log p(n | \textrm{LSTM})=$\textbf{\num{-40.89}} &  \textbf{\num{10}} \\
                                  \makebox[0pt][l]{$s$: }\hphantom{$n$: }Many of the years I did done so. &  $\log p(\mathmakebox[0pt][l]{s}\hphantom{n} | \textrm{ELECTRA})=$\textbf{\num{-35.77}} &              $\log p(\mathmakebox[0pt][l]{s}\hphantom{n} | \textrm{LSTM})=$\num{-46.25} &            \num{0} \\\midrule
                             $n$: There were no guns in the Bronze Age. &              $\log p(n | \textrm{BERT})=$\num{-48.48} &  $\log p(n | \textrm{ELECTRA})=$\textbf{\num{-30.40}} &  \textbf{\num{10}} \\
                          \makebox[0pt][l]{$s$: }\hphantom{$n$: }There is rich finds from the Bronze Age. &     $\log p(\mathmakebox[0pt][l]{s}\hphantom{n} | \textrm{BERT})=$\textbf{\num{-48.46}} &           $\log p(\mathmakebox[0pt][l]{s}\hphantom{n} | \textrm{ELECTRA})=$\num{-44.34} &            \num{0} \\\midrule
                             $n$: You did a great job on cleaning them. &               $\log p(n | \textrm{XLM})=$\num{-40.38} &      $\log p(n | \textrm{RNN})=$\textbf{\num{-43.47}} &  \textbf{\num{10}} \\
                                     \makebox[0pt][l]{$s$: }\hphantom{$n$: }She did a great job at do me. &      $\log p(\mathmakebox[0pt][l]{s}\hphantom{n} | \textrm{XLM})=$\textbf{\num{-39.89}} &               $\log p(\mathmakebox[0pt][l]{s}\hphantom{n} | \textrm{RNN})=$\num{-61.03} &            \num{0} \\\midrule
                        $n$: This logic has always seemed flawed to me. &              $\log p(n | \textrm{LSTM})=$\num{-39.77} &      $\log p(n | \textrm{RNN})=$\textbf{\num{-45.92}} &  \textbf{\num{10}} \\
                   \makebox[0pt][l]{$s$: }\hphantom{$n$: }His cell has always seemed instinctively to me. &     $\log p(\mathmakebox[0pt][l]{s}\hphantom{n} | \textrm{LSTM})=$\textbf{\num{-38.89}} &               $\log p(\mathmakebox[0pt][l]{s}\hphantom{n} | \textrm{RNN})=$\num{-62.81} &            \num{0} \\\midrule
                          \makebox[0pt][l]{$s$: }\hphantom{$n$: }Stand near the cafe and sip your coffee. &               $\log p(\mathmakebox[0pt][l]{s}\hphantom{n} | \textrm{RNN})=$\num{-65.55} &  $\log p(\mathmakebox[0pt][l]{s}\hphantom{n} | \textrm{ELECTRA})=$\textbf{\num{-34.46}} &  \textbf{\num{10}} \\
                             $n$: Sit at the front and break your neck. &      $\log p(n | \textrm{RNN})=$\textbf{\num{-44.18}} &           $\log p(n | \textrm{ELECTRA})=$\num{-34.65} &            \num{0} \\\midrule
                              $n$: Most of my jobs have been like this. &            $\log p(n | \textrm{3-gram})=$\num{-80.72} &     $\log p(n | \textrm{LSTM})=$\textbf{\num{-35.07}} &  \textbf{\num{10}} \\
                          \makebox[0pt][l]{$s$: }\hphantom{$n$: }One of my boyfriend have been like this. &   $\log p(\mathmakebox[0pt][l]{s}\hphantom{n} | \textrm{3-gram})=$\textbf{\num{-80.63}} &              $\log p(\mathmakebox[0pt][l]{s}\hphantom{n} | \textrm{LSTM})=$\num{-41.44} &            \num{0} \\\midrule
                   $n$: They even mentioned that I offer white flowers. &           $\log p(n | \textrm{2-gram})=$\num{-113.38} &     $\log p(n | \textrm{BERT})=$\textbf{\num{-62.81}} &  \textbf{\num{10}} \\
 \makebox[0pt][l]{$s$: }\hphantom{$n$: }But even fancied that would logically contradictory philosophies. &  $\log p(\mathmakebox[0pt][l]{s}\hphantom{n} | \textrm{2-gram})=$\textbf{\num{-113.24}} &             $\log p(\mathmakebox[0pt][l]{s}\hphantom{n} | \textrm{BERT})=$\num{-117.98} &            \num{0} \\
\bottomrule
        \end{tabularx}
        \end{adjustbox}
        \scriptsize
  \caption{\textbf{Examples of pairs of synthetic and natural sentences that maximally contributed to each model's prediction error.} For each model (double row, ``model 1''), the table shows results for two sentences on which the model failed severely. In each case, the failing model 1 prefers synthetic sentence $s$ (higher log probability bolded), while the model it was pitted against (``model 2'') and all 10 human subjects presented with that sentence pair prefer natural sentence $n$. (When more than one sentence pair induced an equal maximal error in a model, the example included in the table was chosen at random.) \label{tab:ext_table_1_natural_vs_synthetic_sentence_pairs}}
\end{table}

\begin{table}[t]
\centering
\scriptsize
\begin{adjustbox}{center}
\begin{tabularx}{18cm}{lllc}
\toprule
                                                                                                     sentence &                                     pseudo-log-likelihood (PLL) &                               approximate log probability &   \# human choices \\
\midrule
                                                                 $s_1$: I found so many in things and called. &              $\log p(s_1 | \textrm{BERT (PLL)})=$\num{-55.14} &     $\log p(s_1 | \textrm{BERT})=$\textbf{\num{-55.89}} &  \textbf{\num{30}} \\
                             $s_2$: Khrushchev schizophrenic so far\\\hphantom{$s_2$: }disproportionately goldfish fished alone. &     $\log p(s_2 | \textrm{BERT (PLL)})=$\textbf{\num{-22.84}} &             $\log p(s_2 | \textrm{BERT})=$\num{-162.31} &            \num{0} \\\midrule
                                                                   $s_1$: Figures out if you are on the lead. &              $\log p(s_1 | \textrm{BERT (PLL)})=$\num{-38.11} &     $\log p(s_1 | \textrm{BERT})=$\textbf{\num{-51.27}} &  \textbf{\num{30}} \\
 $s_2$: Neighbours unsatisfactory indistinguishable\\\hphantom{$s_2$: }misinterpreting schizophrenic on homecoming\\\hphantom{$s_2$: }cheerleading. &     $\log p(s_2 | \textrm{BERT (PLL)})=$\textbf{\num{-16.43}} &             $\log p(s_2 | \textrm{BERT})=$\num{-258.91} &            \num{0} \\\midrule
                                                                    $s_1$: I just say this and not the point. &           $\log p(s_1 | \textrm{ELECTRA (PLL)})=$\num{-34.41} &  $\log p(s_1 | \textrm{ELECTRA})=$\textbf{\num{-33.80}} &  \textbf{\num{30}} \\
           $s_2$: Glastonbury reliably mobilize disenfranchised \\\hphantom{$s_2$: }homosexuals underestimate unhealthy skeptics. &  $\log p(s_2 | \textrm{ELECTRA (PLL)})=$\textbf{\num{-11.81}} &          $\log p(s_2 | \textrm{ELECTRA})=$\num{-162.62} &            \num{0} \\\midrule
                                                            $s_1$: And diplomacy is more people to the place. &           $\log p(s_1 | \textrm{ELECTRA (PLL)})=$\num{-62.81} &  $\log p(s_1 | \textrm{ELECTRA})=$\textbf{\num{-47.33}} &  \textbf{\num{30}} \\
         $s_2$: Brezhnev ingenuity disembarking Acapulco\\\hphantom{$s_2$: }methamphetamine arthropods unaccompanied\\\hphantom{$s_2$: }Khrushchev. &  $\log p(s_2 | \textrm{ELECTRA (PLL)})=$\textbf{\num{-34.00}} &          $\log p(s_2 | \textrm{ELECTRA})=$\num{-230.97} &            \num{0} \\\midrule
                                                           $s_1$: Sometimes what looks and feels real to you. &           $\log p(s_1 | \textrm{RoBERTa (PLL)})=$\num{-36.58} &  $\log p(s_1 | \textrm{RoBERTa})=$\textbf{\num{-51.61}} &  \textbf{\num{30}} \\
                                        $s_2$: Buying something breathes or crawls\\\hphantom{$s_2$: }aesthetically to decorate. &   $\log p(s_2 | \textrm{RoBERTa (PLL)})=$\textbf{\num{-9.78}} &          $\log p(s_2 | \textrm{RoBERTa})=$\num{-110.27} &            \num{0} \\\midrule
                                                   $s_1$: In most other high priority packages were affected. &           $\log p(s_1 | \textrm{RoBERTa (PLL)})=$\num{-71.13} &  $\log p(s_1 | \textrm{RoBERTa})=$\textbf{\num{-61.60}} &  \textbf{\num{30}} \\
                         $s_2$: Stravinsky cupboard nanny contented burglar \\\hphantom{$s_2$: }babysitting unsupervised bathtub. &  $\log p(s_2 | \textrm{RoBERTa (PLL)})=$\textbf{\num{-21.86}} &          $\log p(s_2 | \textrm{RoBERTa})=$\num{-164.70} &            \num{0} \\
\bottomrule
\end{tabularx}
\end{adjustbox}
\scriptsize
\caption{\textbf{Examples of controversial synthetic-sentence pairs that maximally contributed to the prediction error of bidirectional transformers using pseudo-log-likelihood (PLL).} For each bidirectional model, the table displays two sentence pairs on which the model failed severely when its prediction was based on pseudo-log-likelihood (PLL) estimates \cite{wang-cho-2019-bert}. In each of these sentence pairs, the PLL estimate favors sentence $s_2$ (higher PLL bolded), while the approximate log-likelihood estimate and most of the human subjects presented with that sentence pair preferred sentence $s_1$. (When more than one sentence pair induced an equal maximal error in a model, the example included in the table was chosen at random.) \textbf{Sentences with long, multi-token words (e.g., ``methamphetamine'') have high PLL estimates since each of their tokens is well predicted by the others tokens. And yet, the entire sentence is improbable according to human judgments and approximate log-probability estimates based on proper conditional probability chains.}\label{tab:ext_table_2_synthetic_controversial_sentence_pairs_PLL_followup}}
\end{table}

\FloatBarrier
\newgeometry{left=1in, top=1in, right=1in, bottom=1in}
\clearpage
\beginsupplement
\section{Supplementary Methods}
\subsection{Language models\label{sup_methods_language_models}}
{\bf N-gram models.} N-gram models \cite{shannon1948mathematical}, the simplest language model class, are trained by counting the number of occurrences of all unique phrases of length N words in large text corpora. N-gram models make predictions about upcoming words by using empirical conditional probabilities in the training corpus. We tested both 2-gram and 3-gram variants. In 2-gram models, all unique two-word phrases are counted, and each upcoming word probability (probability of $w_2$ conditioned on previous word $w_1$) is determined by dividing the count of 2-gram $w_1,w_2$ by the count of unigram (word) $w_1$. In 3-gram models, all unique three-word phrases are counted, and upcoming word probabilities (probability of $w_3$ conditioned on previous words $w_1$ and $w_2$) are determined by dividing the count of 3-gram $w_1,w_2,w_3$ by the count of 2-gram $w_1,w_2$. In both such models, sentence probabilities can be computed as the product of all unidirectional word transition probabilities in a given sentence. We trained both the 2-gram and 3-gram models on a large corpus composed of text from four sources: 1. public comments from the social media website Reddit (\url{reddit.com}) acquired using the public API at \url{pushshift.io}, 2. articles from Wikipedia, 3. English books and poetry available for free at Project Gutenberg (\url{gutenberg.org}), and 4. articles compiled in the American Local News Corpus \cite{irvine-2014-american}. The n-gram probability estimates were regularized by means of Kneser-Ney smoothing \cite{kneser1995improved}.

{\bf Recurrent neural network models.} We also tested two recurrent neural network models, including a simple recurrent neural network (RNN) \cite{rumelhart1986learning} and a more complex long short-term memory recurrent neural network (LSTM) \cite{hochreiter1997long}. We trained both of these models on a next word prediction task using the same corpus used to train the n-gram models. Both the RNN and LSTM had a 256-feature embedding size and a 512-feature hidden state size, and were trained over 100 independent batches of text for 50 epochs with a learning rate of .002. Both models' training sets were tokenized into individual words and consisted of a vocabulary of \num{94607} unique tokens.

{\bf Transformer models.} Similar to RNNs, transformers are designed to make predictions about sequential inputs. However, transformers do not use a recurrent architecture, and have a number of more complex architectural features. For example, unlike the fixed token embeddings in classic RNNs, transformers utilize context-dependent embeddings that vary depending on a token's position. Most transformers also contain multiple attention heads in each layer of the model, which can help direct the model to relevant tokens in complex ways. We tested five models with varying architectures and training procedures, including BERT \cite{devlin2019bert}, RoBERTa \cite{liu2019roberta}, XLM \cite{Conneau2019Cross}, ELECTRA \cite{clark2020electra}, and GPT-2 \cite{radford2019language}.

\begin{itemize}

\item We used the large version of BERT (bi-directonal encoder representations from transformers), containing 24 encoding layers, \num{1024} hidden units in the feedforward network element of the model, and 16 attention heads. BERT is a bi-directional model trained to perform two different tasks: 1. a masked language modeling (MLM) task, in which 15 percent of tokens are replaced with a special [MASK] token and BERT must predict the masked word, and 2. next sentence prediction (NSP), in which BERT aims to predict the upcoming sentence in the training corpus given the current sentence.

\item RoBERTa is also a bi-directional model that uses the same architecture as BERT. However, RoBERTa was trained on exclusively the masked word prediction task (and not next sentence prediction), and used a different optimization procedure (including longer training on a larger dataset). This makes empirical comparisons between BERT and RoBERTa particularly interesting, because they differ only in training procedure and not architecture. 

\item XLM is a cross-lingual bi-directional model which, too, shares BERT's original architecture. XLM is trained on three different tasks: 1. the same MLM task used in both BERT and RoBERTa, 2. a causal language modeling task where upcoming words are predicted from left to right, and 3. a translation modeling task. On this task, each training example consists of the same text in two languages, and the model performs a masked language modeling task using context from one language to predict tokens of another. Such a task can help the XLM model become robust to idiosyncrasies of one particular language that may not convey much linguistic information. 

\item The ELECTRA model uses a training approach that involves two transformer models: a generator and a discriminator. While the generator performs a masked language modeling task similar to other transformers, the discriminator simultaneously tries to figure out which masked tokens were replaced by the generator. This task may be more efficient than pure masked token prediction, because it uses information from all input tokens rather than only the masked subset.

\item GPT-2, the second iteration of GPT OpenAI's GPT model, is the only unidirectional transformer model that we tested. We used the pretrained GPT-2-xl version, with 48 encoding layers and 25 attention heads in each layer. Because GPT-2 is unidirectional it was trained only on the causal language modeling task, in which tokens are predicted from left to right. 
\end{itemize}

\subsection{Selection of controversial natural-sentence pairs\label{sup_methods_selection}}
We evaluated \num{231725} eight-word sentences sampled from Reddit. Reddit comments were scraped from across the entire website and all unique eight-word sentences were saved. These sentences were subsequently filtered to exclude blatant spelling errors, inappropriate language, and individual words that were not included in the corpus used to train the n-gram and recurrent neural network models in our experiment. 

We estimated $\log p(s \mid m)$ for each natural sentence $s$ and each model $m$ as described above. We then rank-transformed the sentence probabilities separately for each model, assigning the fractional rank $r(s \mid m)=0$ to the least probable sentence according to model $m$ and \mbox{$r(s \mid m)=1$} to the most probable one. This step eliminated differences between models in terms of probability calibration. 

Next, we aimed to filter this corpus for controversial sentences. To prune the candidate sentences, we eliminated any sentence $s$ for which no pair of models $m_1$, $m_2$ held \mbox{$(r(s \mid m_1)<0.5)$} and \mbox{$(r(s \mid m_2)\geq 0.5)$}, where $r(s \mid m_1)$ is the fractional rank assigned for sentence $s$ by model $m$. This step ensured that all of the remaining sentences had a below-median probability according to one model and above-median probability according to another, for at least one pair of models. We also excluded sentences in which any word (except for prepositions) appeared more than once. 
After this pruning, \num{85749} candidate sentences remained, from which $ \binom{85749}{2} \approx 3.67 \times 10^{9}$ possible sentence pairs can be formed.

We aimed to select 360 controversial sentence pairs, devoting 10 sentence pairs to each of the 36 models pairs. First, we defined two 360-long integer vectors $\bm{m}^1$ and $\bm{m}^2$, specifying for each of the 360 yet unselected sentence pairs which model pair they contrast. We then selected 360 sentence pairs $\big(s^1_1,s^2_1\big),\big(s^1_2,s^2_2\big)...,\big(s^1_{360},s^2_{360}\big)$ by solving the following minimization problem:
\begin{subequations}%
\label{eq:natural_controversial_opt}
\begin{alignat}{3}
\{({s_j^1}^*,{s_j^2}^*) \mid j=1,2,..360\}=\argmin_{\bm{s^1},\bm{s^2}} \quad & \sum_j \big( r(s^1_j|m^1_j)+r(s^2_j|m^2_j) \big) \tag{\ref*{eq:natural_controversial_opt}}\\
\textrm{subject to} \quad & \forall_j r(s^1_j|m^2_j) \geq 0.5\\
&\forall_j r(s^2_j|m^1_j)\geq 0.5\\
&\textrm{All 720 sentences are unique.}
\end{alignat}
\end{subequations}

To achieve this, we used integer linear programming (ILP) as implemented by Gurobi \cite{gurobi}. We represented sentence allocation as a sparse binary tensor $\mathrm{S}$ of dimensions \mbox{85,749 $\times$ 360 $\times$ 2} (sentences, trials, pair members) and the fractional sentence probabilities ranks as a matrix $\mathrm{R}$ of dimensions 85,749 $\times$ 9 (sentences, models). This enabled us to express and solve the selection problem in Eq. \ref{eq:natural_controversial_opt} as a standard ILP problem:
\newcommand{\matr}[1]{\bm{#1}} %
\begin{subequations}
\label{eq:natural_controversial_opt_ilp}
\begin{alignat}{3}
\matr{S}^*=\argmin_{\matr{S}} \sum_{i,j}  \quad & \matr{S}_{i,j,1} \matr{R}_{i,m^1_j} + \matr{S}_{i,j,2} \matr{R}_{i,m^2_j}\tag{\ref*{eq:natural_controversial_opt_ilp}}\\
\textrm{subject to} \quad & \matr{S}_{i,j,1} \matr{R}_{i,m^2_j} \geq 0.5\\
& \matr{S}_{i,j,2} \matr{R}_{i,m^1_j} \geq 0.5\\
& \forall_i \sum_{j,k} \matr{S}_{i,j,k}\leq1 \textrm{ (each sentence $i$ is used only once in the experiment)}\\
& \mkern-5mu \begin{rcases}
\forall_j \sum_{i} \matr{S}_{i,j,1}=1\\ 
\forall_j \sum_{i} \matr{S}_{i,j,2}=1\\
\end{rcases}
\textrm{(each trial $j$ is allocated exactly one sentence pair)}\\
&\matr{S} \textrm { is binary}
\end{alignat}
\end{subequations}

\subsection{Evaluation of model-human consistency: Correlating model log-probability ratios to human Likert ratings\label{sup_methods_ordinal}}
For every model $m$ and experimental trial $i$, we evaluated the log probability ratio for the trial's two sentences: 
\begin{equation}
LR(s_i^1, s_i^2 \mid m)=\log \frac{p(s_i^2 \mid m)}{p(s_i^1 \mid m)}
\label{eq:log_ratio}
\end{equation}

The human Likert ratings were recoded to be symmetrical around zero, mapping the six ratings appearing in Extended Data Fig.~1 to $(-2.5,-1.5,-0.5,+0.5,+1.5,+2.5)$. We then sought to correlate the model log-ratios and with the zero-centered human Likert ratings, quantifying how well the model log-ratios were associated with human sentence-likeliness judgments. To allow for an ordinal (not necessarily linear) association between the log-ratios and Likert ratings, we rank-transformed both measures (ranking within each model or each human) while retaining the sign of the values.

\noindent For each participant $h$:
\begin{equation}
r(s_i^1, s_i^2 \mid h)=\sign(y_0(s_i^1, s_i^2 \mid h)) \cdot R\big(\left| y_0(s_i^1, s_i^2 \mid h) \right|\big),
\label{eq:signed_rank_human}
\end{equation}
where $y_0(s_i^1, s_i^2 \mid h))$ is the zero-centered Likert rating provided by subject $h$ for trial $i$ and $R(\cdot)$ is rank transform using random tie-breaking.

\noindent For each model $m$:
\begin{equation}
r(s_i^1, s_i^2 \mid m)=\sign(LR(s_i^1, s_i^2 \mid m)) \cdot R\big(\left| LR(s_i^1, s_i^2 \mid m) \right|\big),
\label{eq:signed_rank_model}
\end{equation}

A valid correlation measure of the model ranks and human ranks must be invariant to whether one sentence was presented on the left ($s_1$) and the other on the right ($s_2$), or vice versa. Changing the sentence order within a trial would flip the signs of both the log-ratio and the zero-centered Likert rating. Therefore, the required correlation measure must be invariant to such coordinated sign flips, but not to flipping the sign of just one of the measures. Since cosine similarity maintains such invariance, we introduced \emph{signed-rank cosine similarity}, an ordinal analog of cosine similarity, substituting the raw data points for signed ranks (as defined in Eq.~\ref{eq:signed_rank_human}-\ref{eq:signed_rank_model}):
\begin{equation}
\label{eq:signed-rank_cosine_similarity}
S_{CSR} = \frac{\sum_i r(s_i^1, s_i^2 \mid m) r(s_i^1, s_i^2 \mid h)}{\sqrt{\sum_i r(s_i^1, s_i^2 \mid m)^2}\sqrt{\sum_i r(s_i^1, s_i^2 \mid h)^2}}.
\end{equation}

To eliminate the noise contributed by random tie-breaking, we used a closed-form expression of the expected value of Eq.~\ref{eq:signed-rank_cosine_similarity} over different random tie-breaking draws:
\begin{equation}
\label{eq:ev_signed-rank_cosine_similarity}
\mathbb{E}(S_{CSR}) =  \frac{\sum_i \mathbb{E}\big( r(s_i^1, s_i^2 \mid m)\big) \mathbb{E}\big( r(s_i^1, s_i^2 \mid h)\big)}{\sqrt{\sum_{k=1}^n k^2}\sqrt{\sum_{k=1}^n k^2}} = \frac{\sum_i \bar{r}(s_i^1, s_i^2 \mid m) \bar{r}(s_i^1, s_i^2 \mid h)}{\sum_{k=1}^n k^2},
\end{equation}
where $\bar{r}(\cdot)$ denotes signed rank with average-rank assigned to ties instead of random tie-breaking, and $n$ denotes the number of evaluated sentence pairs. The expected value of the product in the numerator is equal to the product of expected values of the factors since the random tie-breaking within each factor is independent. The vector norms (the factors in the denominator) are constant since given no zero ratings, each signed-rank rating vector always includes one of each rank $1$ to $n$ (where $n$ is the number of sentence pairs considered), and the signs are eliminated by squaring. This derivation follows a classical result for Spearman's $\rho$ \cite{Woodbury1940Rank} (see \cite{Schutt2023Statistical}, appendix C, for a modern treatment). We empirically confirmed that averaging $S_{CSR}$ as defined in Eq.~\ref{eq:signed-rank_cosine_similarity} across a large number of random tie-breaking draws converges to $\mathbb{E}(S_{CSR})$ as defined in Eq.~\ref{eq:ev_signed-rank_cosine_similarity}. This latter expression (whose computation requires no actual random tie-breaking) was used to quantify the correlation between each participant and model.

For each participant, the lower bound on the noise ceiling was calculated by replacing the model-derived predictions with an across-participants average of the nine other participants' signed-rank rating vectors. The lower bound plotted in main text Fig.~4 is an across-subject average of this estimate. An upper bound on the noise ceiling was calculated as a dot product between the participant's expected signed-rank rating vector ($\mathbf{\bar{r}}/\sqrt{\sum{k^2}}$) and a normalized, across-participants average of the expected signed-rank rating vectors of all 10 participants.%

Inference was conducted in the same fashion as that employed for the binarized judgments (Wilcoxon signed-rank tests across the 10 subject groups, controlling for false discovery rate).

\section{Supplementary Results}
\subsection{Randomly sampled natural-sentence pairs fail to adjudicate among models\label{sup_results_randomly_sampled}}
As a baseline, we created 90 pairs of natural sentence pairs by randomly sampling from a corpus of 8-word sentences appearing on Reddit. Evaluating the sentence probabilities assigned to the sentences by the different models, we found that models tended to agree on which of the two sentences was more probable (Extended Data Fig.~2). The between-model agreement rate ranged from \num{75.6}\% of the sentence pairs for GPT-2 vs. RNN to \num{93.3}\% for GPT-2 vs. RoBERTa, with an average agreement between models of \num{84.5}\%. Main text Fig.~1a (left-hand panel) provides a detailed graphical depiction of the relationship between sentence probability ranks for one model pair (GPT-2 and RoBERTa).

We divided these 90 pairs into 10 sets of nine sentences and presented each set to a separate group of 10 subjects. To evaluate model-human alignment, we computed the proportion of trials where the model and the participant agreed on which sentence was more probable. All of the nine language models performed above chance (50\% accuracy) in predicting the human choices for the randomly sampled natural sentence pairs (main text Fig.~1a, right-hand panel). Since we presented each group of 10 participants with a unique set of sentence pairs, we could statistically test between-model differences while accounting for both participants and sentence pairs as random factors by means of a simple two-sided Wilcoxon signed-rank test conducted across the 10 participant groups. For the set of randomly sampled natural-sentence pairs, this test yielded no significant prediction accuracy differences between the candidate models (controlling for false discovery rate for all 36 model pairs at \textit{q}~$<$~.05). This result is unsurprising considering the high level of between-model agreement on the sentence probability ranking within each of these sentence pairs.

To obtain an estimate of the noise ceiling \cite{nili_toolbox_2014} (i.e., the best possible prediction accuracy for this dataset), we predicted each participant's choices by the majority vote of the nine other participants who were presented with the same trials. This measurement provided a lower bound on the noise ceiling. Including the participant's own choice in the prediction yields an upper bound, since no set of predictions can be more human-aligned on average given the between-subject variability. For the randomly sampled natural sentences, none of the models were found to be significantly less accurate than the lower bound on the noise ceiling (controlling the false discovery rate for all nine models at \textit{q}~$<$~.05). In other words, the 900 trials of randomly sampled and paired natural sentences provided no statistical evidence that any of the language models are human-inconsistent.

\subsection{Comparing the accuracy of unnormalized and normalized sentence probability estimates\label{sup_results_normalization}}
Previous studies (e.g., \cite{lau_2020_how}) have found that normalizing language model sentence probability estimates by the sentences' token counts can result in greater alignment with human acceptability judgments. While we deliberately used unnormalized sentence log probability when designing the experiment, we evaluated the prediction accuracy of each model under such normalizations through a control analysis.

Rather than predicting human judgments based only on the relative log probabilities of the two sentences, we instead used cross-validated logistic regression to predict human judgments using a combination of unnormalized log probability differences (``LP'') and two measures from Lau and colleagues \cite{lau_2020_how} that incorporate information about the token counts in each sentence. The ``MeanLP'' measure normalized each sentence's log probability by its token count $T$, whereas the ``PenLP'' measure divided each sentence's log probability by a dampened version of its token count, $\big((5+T)/(5+1)\big)^{0.8}$. For models trained on whole words (LSTM, RNN, 3-gram, and 2-gram), we used character count instead of token count.

For each language model $m$, we fitted a separate logistic regression to predict the individual binarized sentence choices across the entire main experiment dataset by weighing the three predictors ``LP,'' ``MeanLP,'' and ``PenLP.'' We did not include an intercept due to the symmetry of the prediction task (the presentation of sentences as sentence 1 or 2 was randomized). We cross-validated the logistic regression's accuracy by leaving out one sentence pair at a time, using data from all conditions of the experiment.

\begin{figure}[ht]
\includegraphics[]{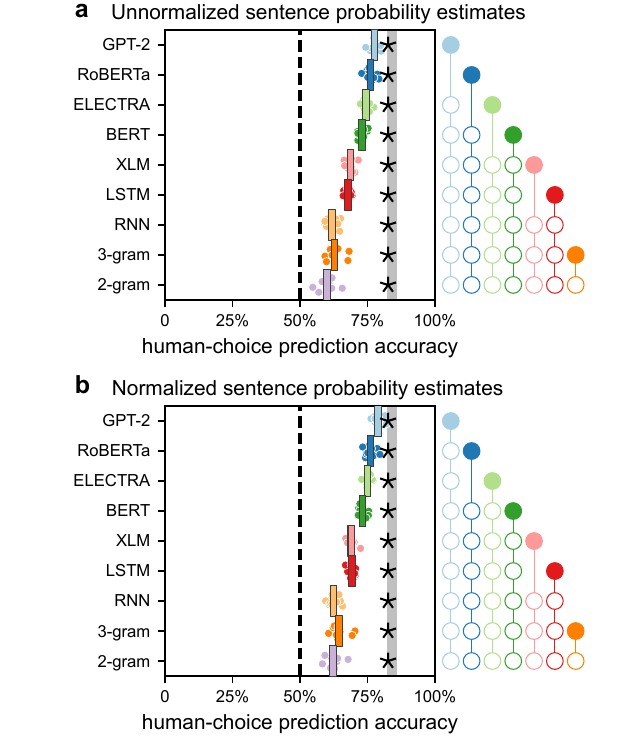}
\caption{\textbf{The predictivity of normalized and unnormalized log-probability measures.} \textbf{(a)} Predicting human judgments from all conditions using only unnormalized log probability differences (equivalent to Fig.~4 in the main text, except using binarized accuracy as a dependent measure). \textbf{(b)} Binarized accuracy of the logistic regression optimally combining LP, MeanLP, and PenLP for each language model. Relative model performance is nearly identical in these two analyses, indicating that tokenization differences across models did not play a large confounding role in our main results.}
\label{fig:s1_token_count_normalized_prediction_accuracy}
\end{figure}

Taking token count into consideration led to minor improvements in the prediction accuracy of most models (an average improvement of $0.95\%$), but this adjustment did not change the hierarchy of the models in terms of their human consistency (Supplementary Fig.~\ref{fig:s1_token_count_normalized_prediction_accuracy}). We hypothesize that the greater disparities between unnormalized and normalized probability measures, as observed by Lau and colleagues \cite{lau_2020_how} compared to those found in our study, may be attributed to their experiment involving sentences of markedly different lengths.

\subsection{Models differ in their sensitivity to low-level linguistic features\label{sup_results_features}}

While the controversial sentences presented in this study were synthesized without consideration for particular linguistic features, we performed a post hoc analysis to explore the contribution of different features to model and human preferences (Supplementary Fig.~\ref{fig:s3_linguistic_feature_analysis}). For each controversial synthetic sentence pair, we computed the average log-transformed word frequency for each sentence (extracted from the publicly available subtlex database \cite{van2014subtlex}). We also computed the average pairwise correlation between semantic GloVe vector representations \cite{pennington2014glove} of all eight words, based on neuroimaging research showing that there are specific neural signatures evoked by dissimilarity in semantic vectors \cite{Frank2017, broderick2018electrophysiological}. We performed paired sample t-tests across sentence pairs between the linguistic feature preferences for models vs. humans, and found that GPT-2, LSTM, RNN, 3-gram, and 2-gram models were significantly more likely (vs. humans) to prefer sentences with low GloVe correlations, while ELECTRA was significantly more likely to prefer high GloVe correlations (controlling the false discovery rate for all nine models at \textit{q}~$<$~.05). For word frequency, the RNN, 3-gram, and 2-gram models were significantly biased (vs. humans) to prefer sentences with low-frequency words, while ELECTRA and XLM showed a significant bias for high-frequency words. These results indicate that even strong models like GPT-2 and ELECTRA can exhibit subtle misalignments with humans in their response to simple linguistic features, when evaluated on sentences synthesized to be controversial.

\captionsetup[figure]{name=Supplementary Fig.}

\pagebreak
\section{Supplementary Figures}

\begin{figure}[ht]
\includegraphics[width=1\linewidth]{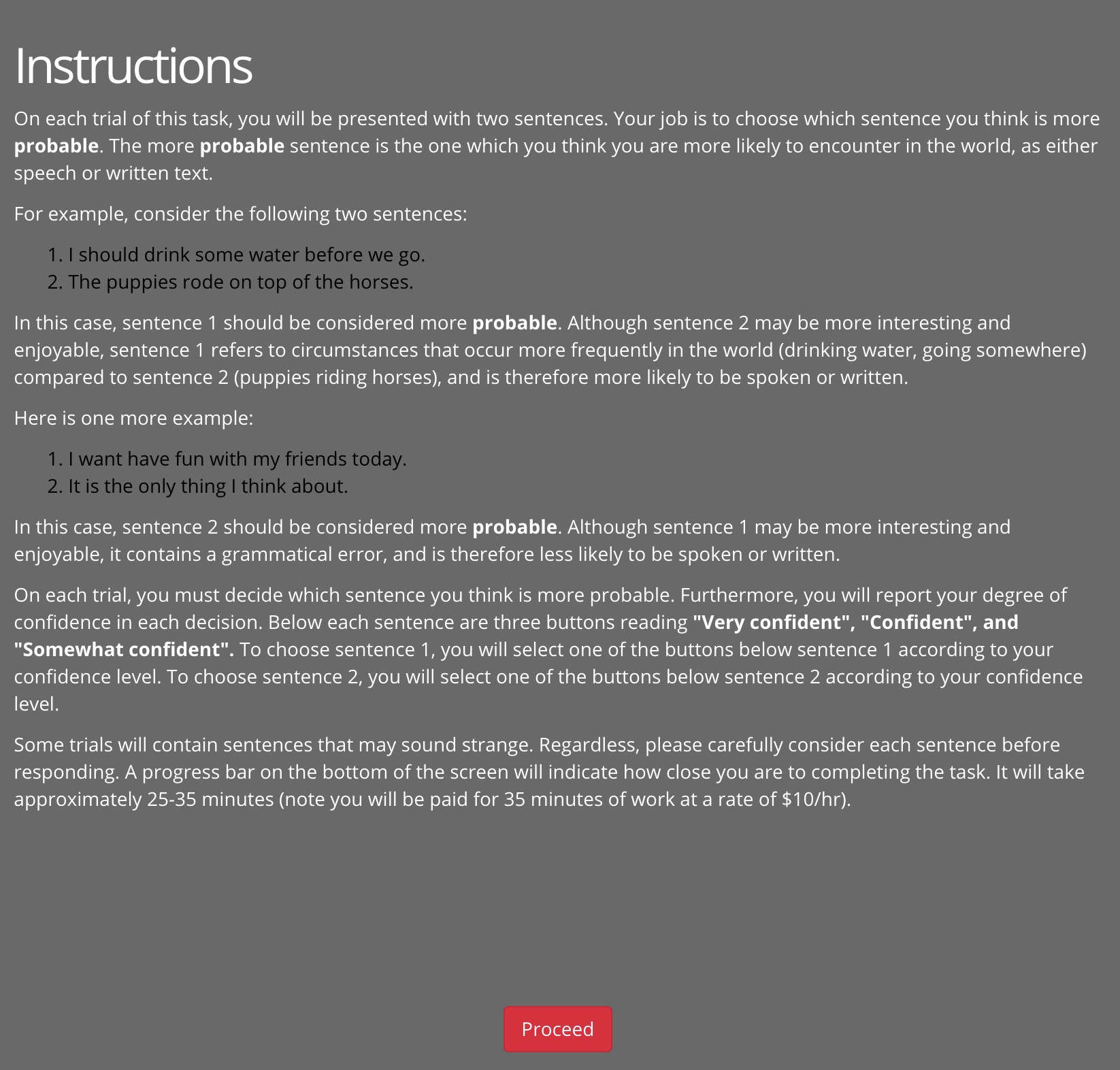}
\caption{The task instructions provided to the participants at the beginning of the experimental session.}
\label{fig:s2_taskinstructions}
\end{figure}

\begin{figure}[htbp]
	\centering
	\includegraphics[width=\textwidth]{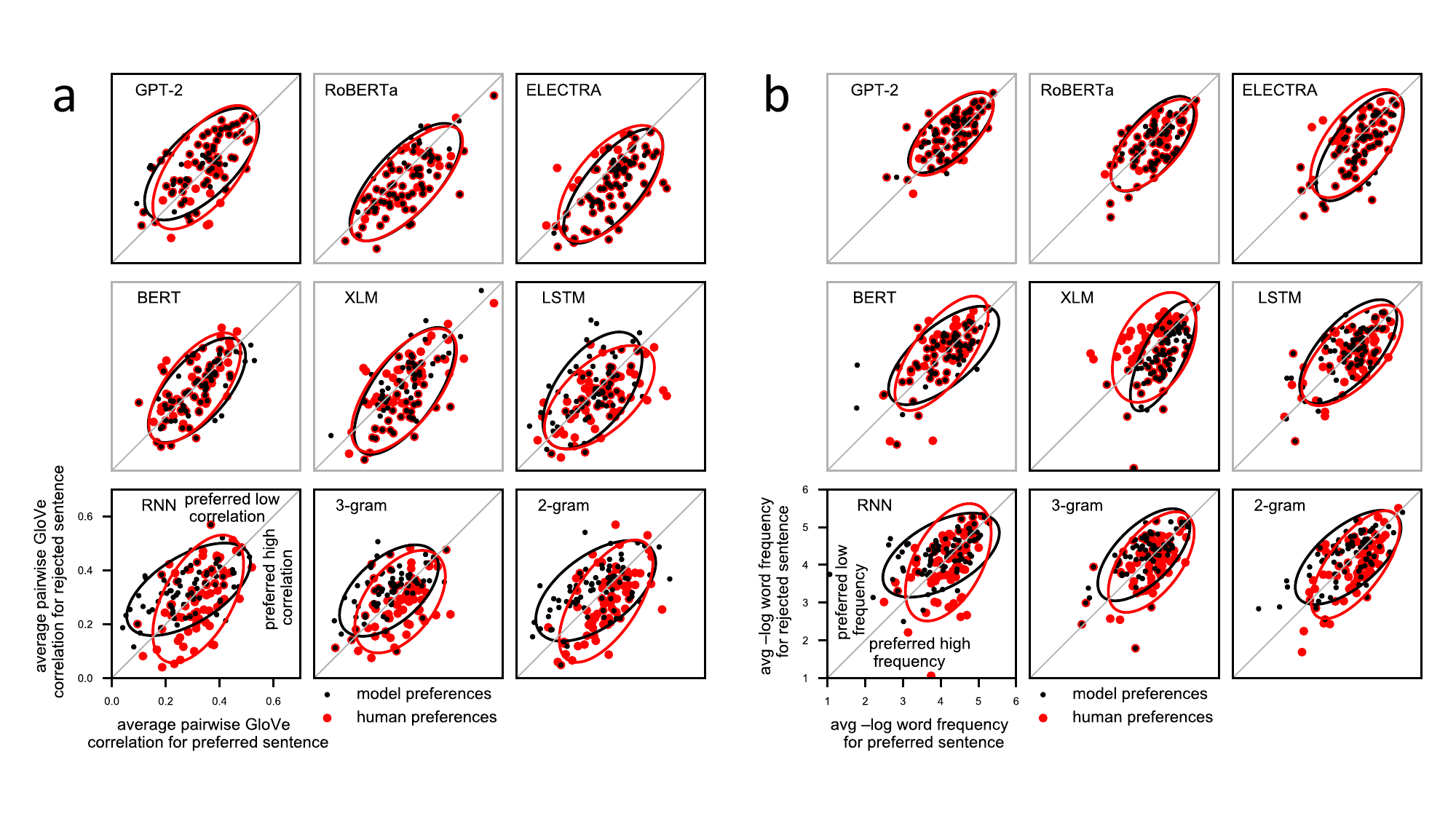}
	\caption{\textbf{Linguistic feature values for synthetic sentence pairs.} \textbf{(a)} GloVe correlation values of the preferred and rejected sentence for each synthetic sentence pair. Each panel depicts preferences for both humans (red) and a specific model (black), for sentence pairs that this model was involved in synthesizing. Black sub-panel outlines indicate significant differences between the preferences of models and humans on that particular set of sentence pairs, according to a paired sample t-test (controlling for false discovery rate across all nine models at \textit{q}~$<$~.05). \textbf{(b)} Same as (a), but for average log-transformed word frequency.}
    \label{fig:s3_linguistic_feature_analysis}
\end{figure}

\FloatBarrier
\captionsetup[table]{name=Supplementary Table}
\section{Supplementary Tables}

    \begin{table}[h]
    \centering
    \begin{tabular}{ccccc}
    \toprule
    model & \shortstack{accepted sentence\\has more tokens} & \shortstack{equal\\token-counts} & \shortstack{rejected sentence\\has more tokens} & p-value \\
    \midrule
    
        GPT-2 & 24 & 13 & 3 & \textbf{$<$0.0001} \\
        
        RoBERTa & 6 & 18 & 16 & 0.0656 \\
        
        ELECTRA & 12 & 21 & 7 & 0.3593 \\
        
        BERT & 4 & 8 & 28 & \textbf{$<$0.0001} \\
        
        XLM & 2 & 16 & 22 & \textbf{$<$0.0001} \\
        
    \bottomrule
    \end{tabular}
    \caption{Token count control analysis. For each transformer model, we considered synthetic controversial sentence pairs where the other targeted model was also a transformer (a total of 40 sentence pairs per model). For each such pair, we evaluated the token count of the synthetic sentence to which the model assigned a higher probability (``accepted sentence'') and the token count of the synthetic sentence to which the model assigned a lower probability (``rejected sentence''). For each model, this table presents the number of sentence pairs in which the accepted sentence had a higher token count, both sentences had an equal number of tokens, and the rejected sentence had a higher token count. We compared the prevalence of higher token counts in accepted and rejected sentences using a binomial test ($H_0: p=0.5$) controlled for False Discovery Rate across five comparisons.\\
    GPT-2 assigned significantly more tokens to accepted sentences, whereas BERT and XLM assigned significantly more tokens to rejected sentences. For RoBeRTa and ELECTRA, no significant difference was found. Note that since the controversial sentences are driven by relative model response properties, a significant difference for a particular model does not necessarily indicate that token count biases the model's sentence probability estimates. For example, GPT-2's apparent preference for sentences with a greater token count might reflect biases of the alternative models pitted against GPT-2. These models might prefer shorter sentences that exhibit undetected grammatical or semantic violations over longer but felicitous sentences.\\
    \textbf{Overall, these results indicate that while certain models' probability estimates might be biased by tokenization, lower sentence probabilities were not systematically confounded by higher token counts.}}
    \label{tab:s1_token_count_analysis}
    \end{table}

\end{document}